\documentclass[preprint, 3p]{elsarticle}

\usepackage{amssymb}
\usepackage{natbib}
\usepackage[T1]{fontenc}
\usepackage{times}
\usepackage{hyperref}
\hypersetup{colorlinks=true,			
			linkcolor=blue}
\usepackage{amsmath, amssymb}
\usepackage{amsthm}
\usepackage{bm, bbm}
\usepackage{mathtools, thmtools}
\usepackage{amsfonts}
\usepackage{multirow}
\usepackage{subcaption}

\newtheorem{thm}{Theorem}
\newtheorem{definition}{Definition}
\newtheorem{lemma}{Lemma}
\newtheorem{corollary}{Corollary}
\newproof{pf}{\textbf{Proof}}
\newtheorem*{note*}{\textbf{Note}}
\newtheorem*{notation*}{\textbf{Notation}}
\newproof{pf1}{\textbf{Proof of Lemma \ref{lemma1}}}
\newproof{pf2}{\textbf{Proof of Lemma \ref{lemma2}}}

\journal{Neural Networks}

\begin{document}

\begin{frontmatter}



\title{Reduced Jeffries-Matusita distance: A Novel Loss Function to Improve Generalization Performance of Deep Classification Models}


\author[]{Mohammad Lashkari \footnotemark[1]{}\footnote[2]{Corresponding author, mohammad.l@aut.ac.ir}}
\author[]{Amin Gheibi \footnotemark[1]{}\footnote[3]{amin.gheibi@aut.ac.ir}}
\address[1]{Department of Computer Science, Amirkabir University of Technology (Tehran polytechnic), Iran}

\begin{abstract}
The generalization performance of deep neural networks in classification tasks is a major concern in machine learning research. Despite widespread techniques used to diminish the over-fitting issue such as data augmentation, pseudo-labeling, regularization, and ensemble learning, this performance still needs to be enhanced with other approaches. In recent years, it has been theoretically demonstrated that the loss function characteristics i.e. its Lipschitzness and maximum value affect the generalization performance of deep neural networks which can be utilized as a guidance to propose novel distance measures. In this paper, by analyzing the  aforementioned characteristics, we introduce a distance called Reduced Jeffries-Matusita as a loss function for training deep classification models to reduce the over-fitting issue. In our experiments, we evaluate the new loss function in two different problems: image classification in computer vision and node classification in the context of graph learning. The results show that the new distance measure stabilizes the training process significantly, enhances the generalization ability, and improves the performance of the models in the Accuracy and F1-score metrics, even if the training set size is small. 
\end{abstract}




\end{frontmatter}


\section{Introduction}\label{intro}
Classification tasks are ubiquitous in many applications such as face recognition, topic modeling, recommender systems, cancer prognosis, etc. One of the main challenges in these tasks is to improve the generalization performance and prevent the over-fitting issue of the output model, designed based on deep neural networks (DNNs). Several methods have been proposed to address this issue \cite{zhang2021bagging, santos2022avoiding, zhang2022graph}. However, the over-fitting of classification models has not been alleviated completely yet. Therefore, new fundamental propositions, derived from theoretical analyses are needed to improve the generalization ability of deep classification models that can be exploited in any subsequent research on deep learning-based classification including those that address the over-fitting issue. In this paper, we focus on loss functions.

Loss functions have a significant role in the training phase of a deep learning model. Theoretically, it has been deduced that the Lipschitz constant and magnitude value of loss functions are related to the generalization ability of DNNs \cite{akbari2021does}. Furthermore, a novel loss function called the Generalized Jeffries-Matusita (GJM) distance was proposed for label distribution learning (LDL) as an alternative to the Kullback-Leibler (KL) divergence to overcome the over-fitting issue of the output models \cite{akbari2021does}. The experiments of \cite{akbari2021does} show that GJM can stabilize the training process of LDL models and increase accuracy. However, their theoretical results are only valid for deep learning models trained by stochastic gradient descent (SGD). Also, according to the definition of GJM, it cannot be used in single-label classification problems. Recently, in the work of \cite{lashkari2023lipschitzness}, we have theoretically demonstrated that the same characteristics of loss functions are also attributed to the generalization performance of the output model obtained by the Adam \cite{kingma2014adam} and AdamW \cite{loshchilov2017decoupled} optimizers which are widespread in the recent years for training DNNs, especially in classification tasks. A well-known loss function to train classification models is Cross-Entropy (CE) which is logarithmic and unbounded. In the present work, using the theoretical results of \cite{akbari2021does, lashkari2023lipschitzness}, we propose a bounded loss function named the Reduced Jeffries-Matusita (RJM) distance which provides better generalization performance for deep classification models, compared to CE. 

In the experiments, we focus on image classification in computer vision and node classification in the graph learning era because there are appropriate benchmarks in these problems challenging the models in the over-fitting issue. We train and test common deep learning architectures in the aforementioned tasks using CE and RJM to analyze the difference between the generalization performance of the output models. For assessing the time cost of RJM, we measure how long the models take to train. The results show how RJM can perform better than CE to improve the generalization performance and increase the accuracy of classification models. 



\section{Related Work}\label{relatedwork}
One of the major criteria to measure the generalization performance is the generalization error, which is defined as the difference between the \textbf{true error} i.e. the expected value of the loss function over the whole input space and the \textbf{training error} i.e. the expected loss value over the training data. A common technique to diminish the generalization error of DNNs is to upper-bound it theoretically. We call these upper bounds \textbf{generalization bounds}.

There are many approaches for deriving generalization bounds such as robustness \cite{zahavy2016ensemble, ren2018learning}, PAC-Bayesian theory \cite{neyshabur2018pac, guan2022fast}, Vapnik-Chervonenkis (VC) dimension \cite{scarselli2018vapnik, basu2018deep, harvey2017nearly}, and uniform stability \cite{bousquet2002stability, shalev2010learnability, hardt2016train, akbari2021does, lashkari2023lipschitzness}. In the robustness approach, a learning algorithm is robust if the slight changes in the training set cannot cause noticeable changes in the training error. It has been proved that the more robust an algorithm is, the less the generalization error of the output model \cite{xu2012robustness}. The PAC-Bayesian theory is an approach for analyzing Bayesian learning algorithms where the hypothesis space has a prior distribution and the output model is a distribution over this space. VC-dimension is a measure to evaluate complexity, flexibility, and the generalization error of classification models whose architecture is a feed-forward or recurrent neural network. Subsequently, it was extended to graph and recursive neural networks \cite{scarselli2018vapnik}. In the previous works conducted under the approaches explained above, generalization bounds in terms of some hyper-parameters, the $L_2$ and the Frobenius norms of the DNNs' weights, and the size of the training set have been derived. However, the particulars of optimization learning algorithms were not considered and the question of which characteristics of loss functions can affect the generalization error of DNNs remains unanswered. Hence, we follow the notion of uniform stability, which is defined for any learning algorithm based on the true error, where the researchers could take steps forward to achieve generalization bounds related to the loss function properties. 

Intuitively, if a learning algorithm is uniformly stable, then the true error of the output model is not sensitive to the noise of training samples. In the work of \cite{bousquet2002stability}, Bousquet and Elisseeff introduced this notion for deterministic learning algorithms. Hardt \textit{et al.} extended it to randomized learning algorithms \cite{hardt2016train}. They found an expected generalization bound directly related to the number of iterations of a learning algorithm. Ali Akbari \textit{et al.} derived a high probability generalization bound for output models obtained by SGD, which is a vanishing function directly related to the Lipschitz constant and the maximum value of a loss function \cite{akbari2021does}. With the same idea, an analogous high probability generalization bound i.e. directly related to the above characteristics has been derived for DNNs trained by Adam or AdamW \cite{lashkari2023lipschitzness}. The theoretical results of \cite{akbari2021does, lashkari2023lipschitzness} can be used as an instruction to choose or create a loss function when the optimizer is SGD or Adam or AdamW. In this paper, by analyzing Lipschitzness and the magnitude value of loss functions, we create RJM and compare it to CE.

\section{Preliminaries}\label{prelim}
Let $X \subseteq \mathbb{R}^{m \times n}$ be the input space of a classification problem and $C \in \mathbb{N}$ be the number of labels. Let $\mathbb{P}_C$ containing probability vectors be the output space and $\mathbbm{1}_C$ containing one-hot encoded vectors be the target space. A deep classification model is $f^{\theta}: X \rightarrow \mathbb{P}_C$ parameterized by $ \theta \in H$ where $H \subset \mathbb{R}^K$ is a bounded set representing the hypothesis space of the model. The format of classification loss functions is $ \ell:\mathbb{P}_C \times \mathbbm{1}_C \rightarrow \mathbb{R}^+$ which compares the predicted probability distribution to the target, having a substantial role in the DNNs training phase. A learning problem is to minimize the true error, defined as $E_{true}(f^{\theta}) \coloneqq \mathbb{E}_{(\mathrm{x}, \mathrm{y})\sim\mathbb{Q}}\left[\ell(f^{\theta}(\mathrm{x}), \mathrm{y})\right]$:

\begin{equation}\label{learningproblem}
	\min_{\theta \in H} E_{true}(f^{\theta}).
\end{equation} 

Since $ \mathbb{Q} $ is unknown, the minimization problem \eqref{learningproblem} is intractable. Therefore, we estimate $ E_{true}(f^{\theta}) $ by  the training error $ E_{train}(f^\theta) \coloneqq \frac1N \sum_{i=1}^{N} \ell(f^\theta(\mathrm{x}_i), \mathrm{y}_i) $ where $ (\mathrm{x}_i, \mathrm{y}_i) $ is a training sample belonging to $ S \in (X \times \mathbbm{1}_C)^N $. To define the generalization error, we also need some definitions and notes.
 
\begin{definition}[Partition]\label{partition}
	Given a training set $S$ of size $ N $, $P_S = \lbrace P_1,P_2,\ldots,P_k\rbrace $ is a partition for $ S $ of size $ k $ if each member of $S$ is in exactly one element of $P_S$ and  $\forall i \ |P_i|  = \frac{N}{k} $.  
\end{definition}

In Definition \ref{partition}, we assumed $ N $ is divisible by $k$. If it is not possible, we repeat a sample enough to make that happen. Each element of $P_S$ represents a mini-batch which is selected in each iteration of the training process to update the parameters.

To train a DNN, we need an iterative optimization algorithm e.g. first-order gradient-based methods. In each iteration of the algorithm, we have to select a mini-batch of the training examples. Suppose that $P_S = \lbrace P_1,P_2,\ldots,P_k\rbrace $ is a partition of the training set $ S $. We use a random sequence $  R=\left(r_1,r_2,\ldots, r_T \right) $ from $\lbrace 1,2,\ldots,k\rbrace $ to specify the selected mini-batch in the arbitrary iteration $1\leq t \leq T$ for updating the model parameters. In the following, $f^\theta_{P_S,R}$ denotes the output model obtained by an optimization algorithm using a partition $ P_S $ and a random sequence $ R $. 

\begin{definition}[Generalization Error]\label{generror}
	 Suppose that $ S $ is a training set and $P_S$ is a partition for it of size $ k $. Consider a random sequence $ R $ from $\lbrace 1,2,\ldots,k \rbrace$. The generalization error of $f^\theta_{P_S,R}$ obtained by an optimization algorithm is defined as 
	 \[
	 GE(f^\theta_{P_S,R}) = E_{true}(f^\theta_{P_S,R}) - E_{train}(f^\theta_{P_S,R}).
	 \]
\end{definition}

\begin{notation*}
	\normalfont
	For simplicity, moving forward, we denote a loss function by $\bm{\ell(\mathrm{\hat{y}}, \mathrm{y})}$ in which the first argument is predicted probability vector and the second argument is the target vector.
\end{notation*}

\begin{definition}[Lipschitzness] \label{losslipdef}
	Let $C \in \mathbb{N}$ be the number of labels of a classification problem. $\ell(\mathrm{\hat{y}}, \mathrm{y})$ is $\gamma$-Lipschitz with regard to its first argument if $\gamma \geq 0$ exists such that $\, \forall \mathrm{\hat{y}_1}, \mathrm{\hat{y}_2} \in \mathbb{P}_C$, we have 
	\begin{equation}\label{lipineq}
	\mathrm{|\ell(\hat{y}_1,y) - \ell(\hat{y}_2,y)|} \leq \gamma \left\| \mathrm{\hat{y}}_1-\mathrm{\hat{y}}_2 \right\|,
	\end{equation}
	where $ \left\|.\right\| $ is the $L_2$-norm. 
\end{definition}

\begin{definition}[Smoothness]
	Given $C\in \mathbb{N}$, $\ell(\mathrm{\hat{y}}, \mathrm{y})$ is $\zeta$-smooth with regard to its first argument if for every $\mathrm{\hat{y}_1}, \mathrm{\hat{y}_2} \in \mathbb{P}_C$, the gradient $\nabla_{\hat{\mathrm{y}}} \ell(\mathrm{\hat{y}}, \mathrm{y})$ holds the inequality \eqref{lipineq} for $\zeta \geq 0$:
	\[
	\mathrm{\left\|\nabla_{\hat{y}_1}, \ell(\hat{y}_1,y) - \nabla_{\hat{y}_2} \ell(\hat{y}_2,y)\right\|} \leq \zeta \left\| \mathrm{\hat{y}}_1-\mathrm{\hat{y}}_2 \right\|.
	\] 
\end{definition}

As mentioned in Section \ref{relatedwork}, we use the notion of uniform stability to link the generalization error with the loss function properties including its Lipschitzness. To define the uniform stability we follow  \cite{hardt2016train,akbari2021does}: 

\begin{definition}[Uniform Stability]\label{uniformstabilitydef}
	Let $S$ and $S'$ be two training sets. Consider two partitions $P_S$ and $P_{S'}$ of size $ k $ which differ in one element. Consider a random sequence $ R $ from $\lbrace 1,2,\ldots,k \rbrace$. Let $f^\theta_{P_S,R}$ and $f^\theta_{P_{S'},R}$ be the output models obtained by an arbitrary optimization algorithm, $A_{opt}$ with the same initialization. Then, $A_{opt}$ is $\beta$-uniform stable with regard to $\ell(\mathrm{\hat{y}}, \mathrm{y})$ if: 
	\[
	\forall S,S' \;\; \sup_{\mathrm{(x,y)}} \mathbb{E}_R \left[ |\ell(f_{P_{S'},R}(\mathrm{x}),\mathrm{y}) - \ell(f_{P_S,R}(\mathrm{x}),\mathrm{y})| \right] \leq \beta.
	\]
\end{definition}

Definition \ref{uniformstabilitydef} states that if an algorithm is uniformly stable with the constant $\beta$, then changing one mini-batch alters the expected value of the loss function over the random sequences at most $\beta$ for any sample of $(\mathrm{x}, \mathrm{y})$. In addition to uniform stability, another factor that affects the generalization error directly is the bounded difference condition \cite{akbari2021does}:
\begin{definition}[BDC]
	Let $k,T \in \mathbb{N}$. $ G :\lbrace 1,2,\ldots,k\rbrace^T \rightarrow \mathbb{R}^+$ satisfies the bounded difference condition (BDC) with the constant $\rho$ if $\, \forall R,R' \in Dom(G)$ which differ in two elements, we have
	\[
	\sup_{R,R'} |G(R')-G(R)| \leq \rho.
	\]
\end{definition}

\section{Generalization Bounds}
\subsection{SGD Optimizer}\label{sgd}
SGD is the simplest gradient-based optimization algorithm that is used to train DNNs i.e. minimizing the training error.
Roughly speaking, in $t$-th iteration of SGD, the current parameters $\theta_{t-1}$ are updated by moving against the direction of the gradient vector with a specified step size:
\begin{equation}\label{sgdstatement}
\theta_t \leftarrow \theta_{t-1} - \eta \nabla_{\theta_{t-1}} \ell(f^{\theta_{t-1}}(\mathrm{x_i}), \mathrm{y}_i),
\end{equation}
where $ \eta $ is the step size which we further denote by the learning rate. If the learning rate is too small, the model remains under-fitted. However, with a large learning rate, the exploding gradient issue occurs. Besides, the performance of the output model is sensitive to the number of iterations. Training the model with too many iterations causes over-fitting even if the learning rate is precisely tuned. 

In the following, we discuss the uniform stability and the generalization error of DNNs trained by SGD. In Theorem \ref{thm1}, the relationship of the Lipschitz constant of a loss function with the uniform stability and BDC is shown \cite{akbari2021does}. Using Theorem \ref{thm1}, a generalization bound for DNNs trained by SGD has been derived \cite{akbari2021does}, clarified in Theorem \ref{thm2}. The proofs are available in the supplementary of \cite{akbari2021does}.

\begin{notation*}
	\normalfont
	As indicated in the statement \eqref{sgdstatement}, SGD uses a specific partition of the training set in which each element has only one sample. In other words, the partition is a set containing the singleton of each training sample. Therefore, we use $f^\theta_{\lbrace S \rbrace,R}$ to denote the output model obtained by SGD in which $ \lbrace S \rbrace  = \lbrace \lbrace (\mathrm{x}, \mathrm{y}) \rbrace \rbrace_{(\mathrm{x}, \mathrm{y}) \in S}$.
\end{notation*}

\begin{thm}\label{thm1}
	\textbf{\cite{akbari2021does}}  Assume SGD runs for $T$ iterations with an annealing learning rate $\eta_t$ to minimize the training error computed on $N$ samples. Let $ \ell(\mathrm{\hat{y}}, \mathrm{y}) $ be $\gamma$-Lipschitz, $\zeta$-smooth, and convex. Then SGD is $\beta$-uniformly stable and for every $ (\mathrm{x},\mathrm{y}) $, $ \ell(f^\theta_{\lbrace S \rbrace,R}(\mathrm{x}),\mathrm{y}) $  satisfies $\rho$-BDC with respect to $R$. Accordingly,
		\[
		\beta \leq \frac{2\gamma^2}{N}\sum_{t=1}^{T} \eta_t, \ \ \  \rho \leq \frac{4\gamma^2}{T}\sum_{t=1}^{T} \eta_t.
		\]
\end{thm}

\begin{thm}\label{thm2}
	\textbf{\cite{akbari2021does}} Consider a loss function $ \ell(\mathrm{\hat{y}}, \mathrm{y}) $ with a maximum value of $L$ which is $\gamma$-Lipschitz, $\zeta$-smooth, and convex. Suppose that SGD is executed for $ T $ iterations with an annealing learning rate $\eta_t$ to obtain $f^\theta_{\lbrace S \rbrace,R}$ by minimizing the training error achieved from $N$ samples. Then we have the following inequality with probability at least $1-\delta$:
	\begin{equation}\label{sgd-gen-bound}
		GE(f^\theta_{\lbrace S \rbrace,R}) \leq 2\gamma^2 \sum_{t=1}^{T} \eta_t \left(2\sqrt{\frac{\log(2/\delta)}{T}} + \sqrt{\frac{2\log(2/\delta)}{N}} + \frac1N \right) + L\sqrt{\frac{\log(2/\delta)}{2N}}. 
	\end{equation}
\end{thm}

\begin{note*}
	\normalfont
	The inequality \eqref{sgd-gen-bound} demonstrates that choosing a loss function with a lower Lipschitz constant and maximum value reduces the generalization error of an output model obtained by SGD, which leads to overcoming the over-fitting issue.
\end{note*}
\subsection{Adam Optimizer}
SGD has two major disadvantages. First, in each iteration, it takes just one sample of the training set to update the parameters. Second, it is sensitive to the norm of the gradient vector which makes the process of learning rate tuning more involved. This can result in the problem vanishing or exploding gradient. To address these issues, adaptive gradient algorithms were introduced \cite{hazan2015beyond, tieleman2012lecture, duchi2011adaptive, kingma2014adam, loshchilov2017decoupled}. One of the most useful of these optimizers is the adaptive moment estimation (Adam). Intuitively, Adam utilizes the first-moment estimate of the gradient for determining the appropriate direction and also uses the second-moment estimate to neutralize the effect of the gradient norm.

Let $ \ell(f^\theta; P)$ be the loss function value, computed on an arbitrary mini-batch $P=\lbrace(\mathrm{x}_i, \mathrm{y}_i)\rbrace_{i=1}^b$:
 \[  
 \ell(f^\theta;P) = \frac{1}{b}\sum_{i = 1}^{b}\ell(f^\theta(\mathrm{x}_i), \mathrm{y}_i),
 \]
where $b$ is the size of $P$ denoting the batch size. In the following, we get to the particulars of Adam. All operators in  \eqref{m_t}-\eqref{adamstatement} are element-wise. Let $m_t$ and $v_t$ be the first and second moment estimates respectively which are defined as: 
 \begin{align}
 m_t &\leftarrow \beta_1 \cdot m_{t-1} + (1-\beta_1) \cdot g(\theta_{t-1}), \label{m_t} \\
 v_t &\leftarrow \beta_2 \cdot v_{t-1} + (1-\beta_2) \cdot g^2(\theta_{t-1}) \label{v_t},
 \end{align}
 where $ \beta_1, \beta_2 \in (0,1), m_0=0, v_0=0 $, and $ g(\theta) = \nabla_\theta \ell(f^\theta; P)$. The bias-corrected versions of $m_t$ and $v_t$ are:
  
 \begin{align}
 	\widehat{m}_t &\leftarrow \frac{m_t}{1-\beta_1^t}, \label{m_hat_t} \\
 	\widehat{v}_t &\leftarrow \frac{v_t}{1-\beta_2^t}. \label{v_hat_t}
 \end{align}
 Adam updates the parameters as
 
 \begin{equation}\label{adamstatement}
 \theta_t \leftarrow \theta_{t-1} - \eta \cdot \frac{\widehat{m}_t}{(\sqrt{\widehat{v}_t}+\epsilon)},
 \end{equation}
 where $ \eta $ is the learning rate and $ \epsilon=10^{-8} $.
 Now, we analyze the generalization error. In Theorem \ref{thm3}, we discuss the uniform stability of Adam. In Theorem \ref{thm4}, a generalization bound for a DNN trained by Adam is argued, which gives us the same information about the effect of loss functions on the generalization error of SGD that we clarified in Subsection \ref{sgd}:
 
 \begin{thm}\label{thm3}
 	\textbf{\cite{lashkari2023lipschitzness}}  Assume Adam runs for $T$ iterations with a learning rate $\eta$ and batch size $b$ to minimize the training error computed on $N$ samples. Consider a convex and $\gamma$-Lipschitz  loss function $ \ell(\mathrm{\hat{y}}, \mathrm{y}) $. Then Adam is $\beta$-uniformly stable and for every $ (\mathrm{x},\mathrm{y}) $, $ \ell(f^\theta_{P_S,R}(\mathrm{x}),\mathrm{y}) $  satisfies $\rho$-BDC with respect to $R$. Additionally, we have
 	\[
 	\beta \leq \frac{2\eta}{c} \cdot \frac{bT\gamma^2}{N}, \ \ \ \rho \leq \frac{8\eta}{c} \cdot \left( \frac{b\gamma}{N} \right)^2,
 	\]
 	where $c \in (0,1)$ is constant.
 \end{thm}
 
\begin{thm}\label{thm4}
	\textbf{\cite{lashkari2023lipschitzness}} Consider a loss function $ \ell(\mathrm{\hat{y}}, \mathrm{y}) $ with a maximum value of $L$ which is $\gamma$-Lipschitz and convex. Suppose that Adam is executed for $ T $ iterations with a learning rate $\eta$ and batch size $b$ to obtain $f^\theta_{P_S,R}$ by minimizing the training error achieved from $N$ samples. Then with probability at least $1-\delta$, we have
	\begin{equation}\label{adam-gen-bound}
	GE(f^\theta_{P_S,R}) \leq \frac{2\eta}{c} \left( 4\left( \frac{b\gamma}{N} \right)^2 \sqrt{T\ log(2/\delta)} + \frac{bT\gamma^2}{N} \left( 1+\sqrt{2N\log(2/\delta)} \right) \right) + L\sqrt{\frac{\log(2/\delta)}{2N}},
	\end{equation}
	where $c \in (0,1)$ is constant.
\end{thm}  

\subsection{AdamW Optimizer}
A common technique to improve the generalization performance of deep learning models is to exploit a regularization parameter in loss functions. This parameter makes the hypothesis space smaller and reduces the over-fitting issue. However, when we train the models by Adam, regularization does not decay the weight of the model parameters \cite{loshchilov2017decoupled}. The idea of AdamW \footnote{Adam with decoupled weight decay (AdamW)} is to remove the regularizer from the loss function and use it in the update statement. Let $\ell^{reg}(f^\theta; P)$ represents the regularized version of $ \ell(f^\theta; P) $:
\begin{equation} \label{regloss}
\ell^{reg}(f^\theta;P) = \ell(f^\theta; P) + \frac{\lambda}{2b}\left\| \theta \right\|^2,
\end{equation}
where $\lambda \in \mathbb{R}^+$ is the regularizer (weight decay) and $ b $ is the batch size.

Using the weight decay as the equation \eqref{regloss} is not effective in Adam \cite{loshchilov2017decoupled}. Therefore, AdamW decouples this parameter and uses $ \ell(f^\theta; P)  $ to compute the gradient same as Adam. It adds the weight decay to the update statement directly. Consider $\widehat{m}_t$ and $\widehat{v}_t$ in the statements \eqref{m_hat_t} and \eqref{v_hat_t}. Given a learning rate $ \eta $ and schedule multiplier $\alpha_t$, the AdamW's update statement is:
\[
\theta_t \leftarrow \theta_{t-1} - \alpha_t \left( \eta \cdot \frac{\widehat{m}_t}{(\sqrt{\widehat{v}_t}+\epsilon)} + \lambda\theta_{t-1} \right).
\] 
Consider a hypothesis space, $ H $. Note that $H$ is bounded. Let 
\[
 \left\| \theta \right\|_{\sup} \vcentcolon =  \sup_{\theta \in H} \left\| \theta \right\|. 
\]
Now, we can state the theorems:

\begin{thm}\label{thm5}
	\textbf{\cite{lashkari2023lipschitzness}}  Assume AdamW runs for $T$ iterations with a learning rate $\eta$, schedule multiplier $\alpha_t$, weight decay $\lambda$, and batch size $b$ to minimize the training error computed on $N$ samples. Consider a convex and $\gamma$-Lipschitz  loss function $ \ell(\mathrm{\hat{y}}, \mathrm{y}) $. Then AdamW is $\beta$-uniformly stable and for every $ (\mathrm{x},\mathrm{y}) $, $ \ell(f^\theta_{P_S,R}(\mathrm{x}),\mathrm{y}) $  satisfies $\rho$-BDC with respect to $R$. Besides, 
	\[
	\beta \leq \frac{2bT}{N} \left( \frac{\eta\gamma^2}{c} + \gamma \lambda \left\| \theta \right\|_{\sup} \right) \sum_{t=1}^{T} \alpha_t, \ \ \ \rho \leq \frac{8b^2}{N^2} \left( \frac{\eta\gamma^2}{c} + \gamma \lambda \left\| \theta \right\|_{\sup} \right) \sum_{t=1}^{T} \alpha_t,
	\]
	where $c \in (0,1)$ is constant.
\end{thm}

\begin{thm}\label{thm6}
	\textbf{\cite{lashkari2023lipschitzness}} Consider a loss function $ \ell(\mathrm{\hat{y}}, \mathrm{y}) $ with a maximum value of $L$ which is $\gamma$-Lipschitz and convex. Suppose that AdamW is executed for $ T $ iterations with a learning rate $\eta$, schedule multiplier $\alpha_t$, weight decay $\lambda$, and batch size $b$ to obtain $f^\theta_{P_S,R}$ by minimizing the training error achieved from $N$ samples. Then with probability at least $1-\delta$, we have
	\[
	GE(f^\theta_{P_S,R}) \leq \frac{2b}{N} \left( \frac{\eta\gamma^2}{c} + \gamma \lambda \left\| \theta \right\|_{\sup} \right) \left( \frac{4b}{N}\sqrt{T\log(2/\delta)} + T\sqrt{2N\log(2/\delta)} \right) \sum_{t=1}^{T} \alpha_t + L\sqrt{\frac{\log(2/\delta)}{2N}},
	\]
	where $c \in (0,1)$ is constant.
\end{thm}  

\section{Loss Functions}
In this section, we analyze the characteristics of classification loss functions, based on the generalization bounds previously mentioned. CE is a logarithmic function that is derived from the maximum log-likelihood estimation to classify samples in a label set of size $C$:
\[
\ell_{CE}(\mathrm{\hat{y}}, \mathrm{y}) = -\sum_{c=1}^{C} y_c\log(\hat{y}_c). 
\]
Our proposed loss function, RJM is a reduced version of Jeffries-Matusita distance \cite{matusita1955decision}: 
\[
\ell_{RJM}(\mathrm{\hat{y}}, \mathrm{y}) = \sum_{c=1}^{C} y_c(1-\sqrt{\hat{y}_c}).
\]
Based on Theorems \ref{thm2}, \ref{thm4}, and \ref{thm6}, we should prove that the Lipschitz constant and maximum value of RJM are less than CE. In Lemma \ref{lemma1}, we prove general properties for classification loss functions. Lemma \ref{lemma2} shows the relationship between the absolute values of a real function and its Lipschitzness. Corollary \ref{coro1} demonstrates the convexity of CE and RJM. Finally, using Lemmas \ref{lemma1} and \ref{lemma2}, we state Theorem \ref{thm7}, showing the relationship between the Lipschitz constant and the maximum value of CE and RJM.

\begin{lemma}\label{lemma1}
	Consider a function $h: (0,1) \rightarrow \mathbb{R}^+$. The identity loss function for  classification problems is defined as:
	\[
	I(\mathrm{\hat{y}}, \mathrm{y}) = \sum_{c=1}^{C} y_c h(\hat{y}_c).
	\]
	If $h(.)$ is $\gamma$-Lipschitz i.e. 
	\begin{equation}\label{hlipdef}
	\forall u, v \in Dom(h) \;\; \left|h(u) - h(v)\right| \leq \gamma \left| u-v \right|,
	\end{equation}
	Then $ I(\mathrm{\hat{y}}, \mathrm{y}) $ is $\gamma\sqrt{C}$-Lipschitz. Furthermore, $ I(\mathrm{\hat{y}}, \mathrm{y}) $ is convex with regard to its first argument if $h(.)$ is convex. 
\end{lemma}

\begin{pf}
	Consider two probability vectors $\mathrm{u}=\left[u_c\right]_{c=1}^{C}$, $\mathrm{v}=\left[v_c\right]_{c=1}^{C}$, and the target vector $ \mathrm{y} \in \mathbbm{1}_C $. According to Definition \ref{losslipdef} we have
	\begin{align}
	\left| I(\mathrm{u}, \mathrm{y}) - I(\mathrm{v},\mathrm{y}) \right| &= \left| \sum_{c=1}^{C} y_c h(u_c) - \sum_{c=1}^{C} y_c h(v_c)\right| \notag \\
	&\leq \sum_{c=1}^{C} y_c \left| h(u_c) - h(v_c)\right| \notag \\
	&\leq \gamma \sum_{c=1}^{C} y_c \left| u_c - v_c \right| \notag \\
	&\leq \gamma\sqrt{C} \left\|\mathrm{u} - \mathrm{v} \right\|  \label{eq28}.
	\end{align}
	In the inequality \eqref{eq28}, Cauchy-Schwartz is applied. 
	
	Now we prove the convexity of $I(\mathrm{\hat{y}}, \mathrm{y})$ subject to the convexity of $ h(.) $. For $ t \in [0,1] $ we have
	\begin{align*}
	I(t\mathrm{u} + (1-t)\mathrm{v}, \mathrm{y}) &= \sum_{c=1}^{C} y_c h(t u_c + (1-t) v_c) \\
	&\leq \sum_{c=1}^{C} y_c(t h(u_c) + (1-t)h(v_c)) \\
	&= t\sum_{c=1}^{C} y_c h(u_c) + (1-t)\sum_{c=1}^{C} y_c h(v_c) \\
	&= t I(\mathrm{u}, \mathrm{y}) + (1-t) I(\mathrm{v}, \mathrm{y}).
	\end{align*}
\end{pf}

\begin{corollary}\label{coro1}
	$ \ell_{CE}(\mathrm{\hat{y}}, \mathrm{y}) $ and  $ \ell_{RJM}(\mathrm{\hat{y}}, \mathrm{y}) $ are convex.
\end{corollary}

\begin{pf}
	According to the definition of $I(\mathrm{\hat{y}}, \mathrm{y})$ in Lemma \ref{lemma1}, we have $h_{CE}(x) = -\log(x)$ and $h_{RJM}(x) = 1-\sqrt{x}$. Note that $\log(x)$ and $\sqrt{x} $ are convex. Thus, the proposition is concluded.
\end{pf}

\begin{note*}
	\normalfont
	In Lemma
	\ref{lemma1}, $h(.)$ should be chosen in such a way that $\lim_{x \rightarrow 1^-}h(x) = 0$. Otherwise, the training process will not proceed properly and by minimizing the loss function the performance of deep learning models may decrease. Both $ h_{CE}(x)$ and $h_{RJM}(x)$ have this property.
\end{note*}

\begin{lemma}\label{lemma2}
	Let $h: (0,1) \rightarrow \mathbb{R}^+$ be a derivative function. If a constant $\gamma$ exists such that 
	\[ 
	\gamma = \sup_x \left| h'(x) \right|, 
	\]
	Then $h(.)$ is $\gamma$-Lipschitz.	
\end{lemma}

\begin{pf}
	This lemma is proved from the Lipschitzness definition of $h(.)$ by the Mean Value Theorem.
\end{pf}

\begin{thm}\label{thm7}
	Let $\gamma_{CE}$, $\gamma_{RJM}$ be the Lipschitz constants of $ \ell_{CE}(\mathrm{\hat{y}}, \mathrm{y}) $ and $ \ell_{RJM}(\mathrm{\hat{y}}, \mathrm{y}) $ respectively. We have
	\begin{equation}\label{thm7-ineq1}
	\gamma_{CE} \leq \gamma_{RJM}.
	\end{equation}
	In addition, RJM is upper-bounded by CE:
	\begin{equation}\label{thm7-ineq2}
	\ell_{RJM}(\mathrm{\hat{y}}, \mathrm{y}) \leq \ell_{CE}(\mathrm{\hat{y}}, \mathrm{y}).
	\end{equation}
\end{thm}

\begin{pf}
We first prove the inequality \eqref{thm7-ineq1}. According to Lemmas \ref{lemma1} and \ref{lemma2}, it is enough to prove that 
\begin{equation} \label{ineq3}
\left| h'_{RJM}(x) \right| \leq \left| h'_{CE}(x) \right|. 
\end{equation}
Simplifying the inequality \eqref{ineq3} gives
\begin{equation}\label{ineq4}
0 < x \leq 2\sqrt{x}.
\end{equation} 	
The inequality \eqref{ineq4} holds for $0 < x \leq 4$. Note that $Dom(h)=(0,1)$. Therefore, the inequality \eqref{ineq4} is satisfied.

Let's prove the second part.  Using Bernoulli's inequality 
$\exp(x) \geq 1+x$, we have
\begin{align}
\ell_{RJM}(\mathrm{\hat{y}}, \mathrm{y}) &= \sum_{c=1}^{C} y_c(1-\sqrt{\hat{y}_c}) \notag \\
&= \sum_{c=1}^{C} y_c(1-\exp(\frac12 \log(\hat{y}_c))) \notag \\
&\leq \sum_{c=1}^{C} y_c(-\frac{1}{2}\log(\hat{y}_c)) \label{ineq5} \\
&\leq \ell_{CE}(\mathrm{\hat{y}}, \mathrm{y}). \notag
\end{align}
By Bernoulli's inequality, we argued the inequality \eqref{ineq5}.
\end{pf}
\begin{corollary}\label{col2}
	Let $f^{\theta,CE}_{P_S,R}$ and $f^{\theta,RJM}_{P_S,R}$ be the output models obtained by SGD or Adam or AdamW using CE and RJM respectively under the same settings for hyper-parameters. Then, the upper bound of $GE(f^{\theta,RJM}_{P_S,R})$ is less than the upper bound of $GE(f^{\theta,CE}_{P_S,R})$.
\end{corollary}
\begin{pf}
	Let $L_{CE}$ and $L_{RJM}$ be the maximum values of CE and RJM respectively. From Theorem \ref{thm7}, we have $ \gamma_{CE} \leq \gamma_{RJM} $ and $ L_{CE} \leq L_{RJM} $. By applying these inequalities to generalization bounds stated in Theorems \ref{thm2}, \ref{thm4} and \ref{thm6}, we argue the preposition.
\end{pf}

\section{Experiments}
In our experimental evaluation, we do not compete with the state-of-the-art classification models. The goal is to show the effect of RJM on the generalization error of DNNs in node and image classification tasks. We train the models by Adam, AdamW, and SGD using CE and RJM under the same settings for hyper-parameters to fairly compare our loss function with the previous one.  
\subsection{Image Classification}
\subsubsection{Problem Formulation}
Consider a label set $\lbrace 1,2,\ldots,C \rbrace$. Let $(\mathrm{x}, \mathrm{y})$ represents a sample of this problem where $ \mathrm{x} $ is the input image and $ \mathrm{y} \in \mathbbm{1}_C$ is the true label. Then, $\hat{\mathrm{y}} = f^\theta(\mathrm{x})$ indicates the output of a deep learning model $f^\theta$ given the input $ \mathrm{x} $ which is a probability vector of size $C$. The predicted label corresponding to $\mathrm{x}$ is the index of the largest element of $ \hat{\mathrm{y}} $.

To obtain $f^\theta$ we choose the ResNet50 \cite{resnet50} and VGG16 \cite{vgg16} architectures. These convolution-based models were pre-trained on large computer vision datasets and provided adequate results in several classification tasks. We train these models using both CE and RJM to show how much RJM can reduce the generalization error of an image classifier in practice.   

\subsubsection{Dataset and Settings}
We utilize the Intel \cite{intelimage2018} dataset containing images of natural scenes around the world. There are $14034$ training samples and $3000$ test samples distributed almost uniformly under $6$ different classes of building, forest, glacier, mountain, sea, and street in various lights and colors of size $150\times150$. To estimate the generalization error properly in each step of the training phase, it is necessary to have an adequate number of validation samples. Hence, by separating $4034$ samples randomly from the training set, we get to the training, validation, and test sets called Intel-Train, Intel-Val, and Intel-Test respectively. We augment the samples by random cropping and flipping horizontally.

We use ResNet50 and VGG16 pre-trained on ImageNet \cite{deng2009imagenet}. The last layer of ResNet50 is replaced by two dense layers of sizes $ 512 $ and $ 6 $, respectively, and the last layer of VGG16 is replaced with one dense layer of size $ 6 $. AutoGrad does not run over the other layers. We use the same random seed to initialize the new parameters. ResNet50 is trained by Adam and AdamW. SGD is used to train VGG16. The batch size and weight decay are set to $64$ and $0.1$ respectively. The models are trained in $20$ epochs. The learning rate value in each epoch is reported in Table \ref{lrtabel} \footnote{In AdamW, scheduling the learning is handled leveraging the scheduling multiplier parameter. However, In our theoretical results for Adam, the learning rate in the training process is fixed, but it does not affect the correctness of our theorems because we can replace $\eta$ by the largest value for the learning rate in the generalization bound \eqref{adam-gen-bound}.}. The parameters $\beta_1$ and $\beta_2$ are set to $0.9$ and $0.999$ respectively which are suggested by \cite{kingma2014adam, loshchilov2017decoupled}.

\begin{table}[h]
	\begin{center}
		\caption{Learning Rate Settings}
		\label{lrtabel}
		\begin{tabular}{|c|c|}
		\hline
		Optimizer & Learning rate\\
		\hline \hline  
		Adam &  $10^{-4} $ in epochs $ 1 $ to $9$; $ 10^{-5} $ in epochs $10$ to $20$\\
		\hline
		AdamW & same as Adam  \\
		\hline 
		SGD & $10^{-3}$ in epochs $ 1 $ to $9$; $2 \times 10^{-4}$ in epochs $10$ to $14$; $4 \times 10^{-5}$ in epochs $15$ to $20$ \\
		\hline
		\end{tabular}
	\end{center}
\end{table}

\subsubsection{Evaluation}
\begin{figure}[h]
	\begin{center}
		\begin{subfigure}[b]{0.32\textwidth}
			\includegraphics[height=3.33cm, width=5cm]{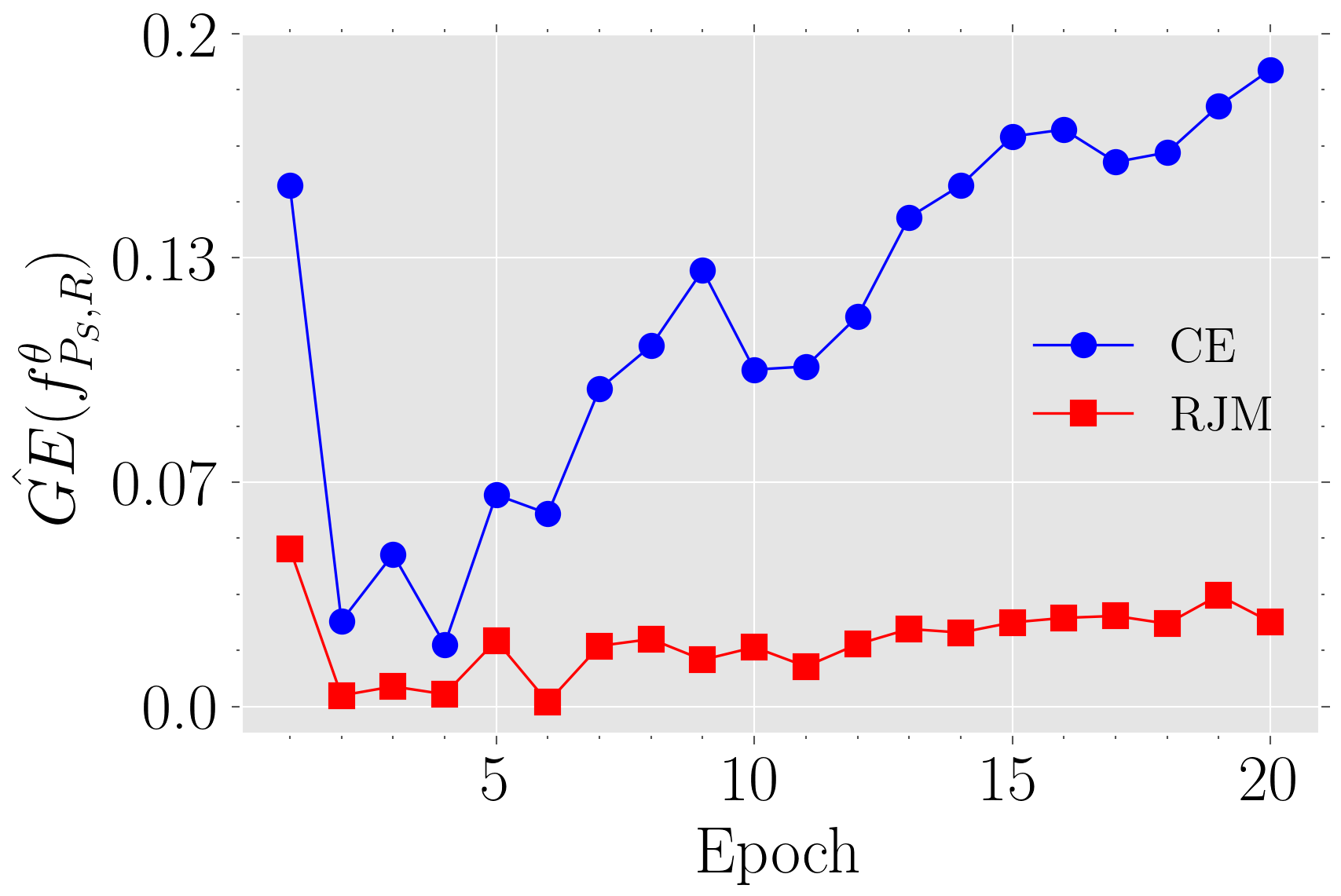}
			\caption{Generalization error estimate}
			\label{GE-ResNet-Adam-CE-RJM}
		\end{subfigure}
		\hfill
		\begin{subfigure}[b]{0.32\textwidth}
			\includegraphics[height=3.33cm, width=5cm]{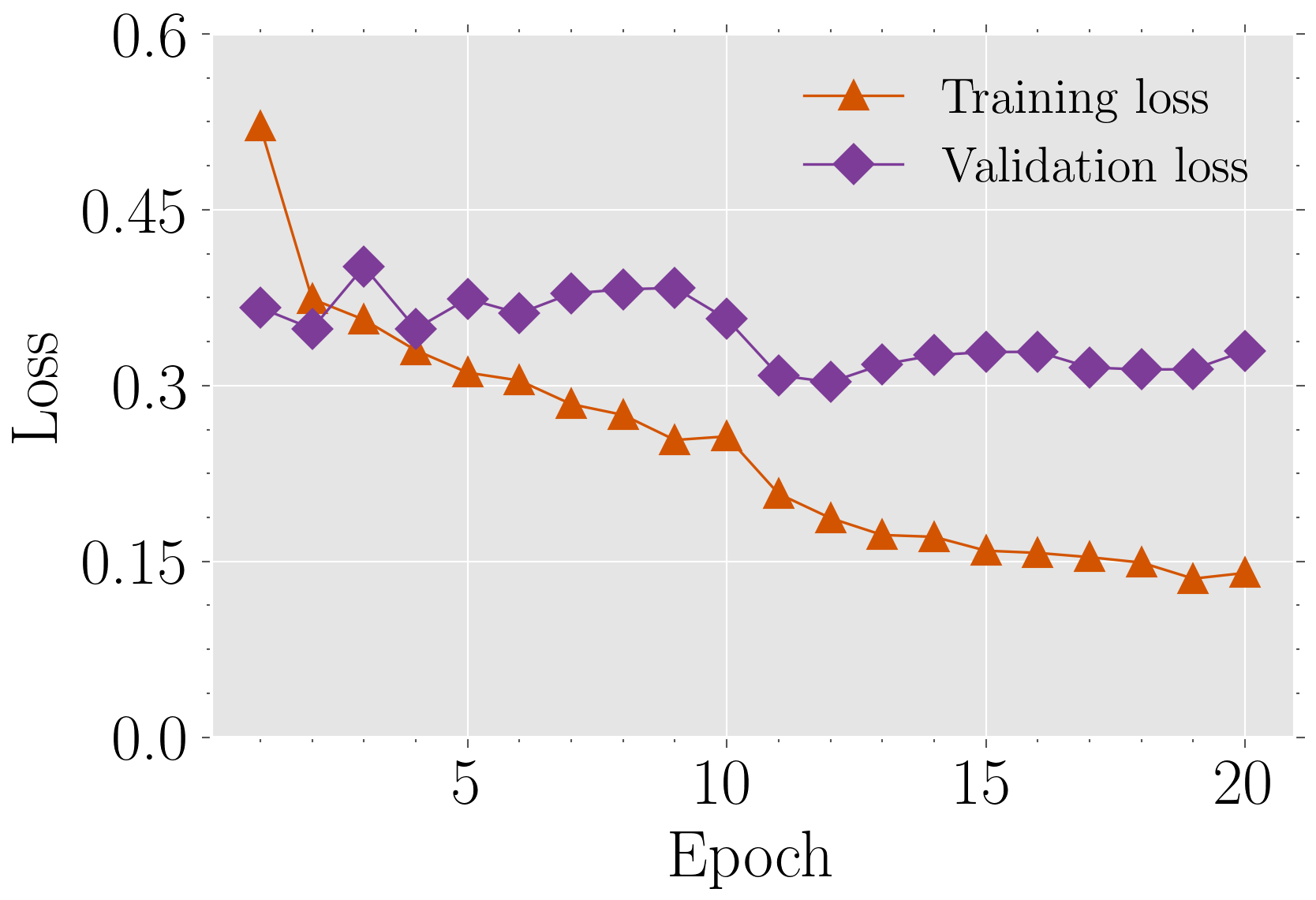}
			\caption{CE on the training and validation sets}
			\label{ResNet-Adam-CE}
		\end{subfigure}
		\hfill
		\begin{subfigure}[b]{0.32\textwidth}
			\includegraphics[height=3.33cm, width=5cm]{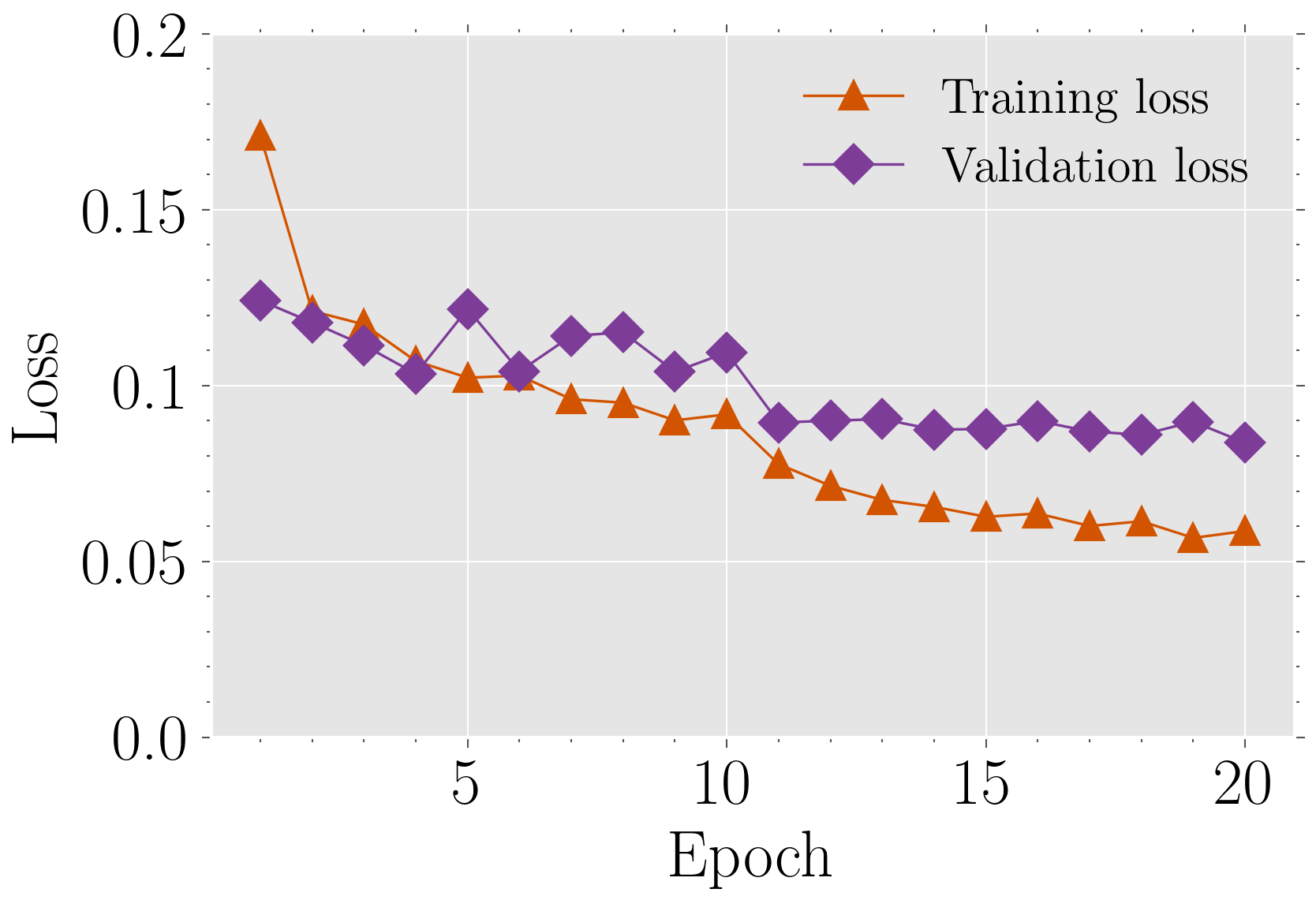}
			\caption{RJM on the training and validation sets}
			\label{ResNet-Adam-RJM}
		\end{subfigure}
		\caption{Evaluation in terms of the generalization error estimate and loss values (Model: ResNet50, Optimizer: Adam)}
	\end{center}
\end{figure}

\begin{figure}[h]
	\begin{center}
		\begin{subfigure}[b]{0.3\textwidth}
			\includegraphics[height=3.33cm, width=5cm]{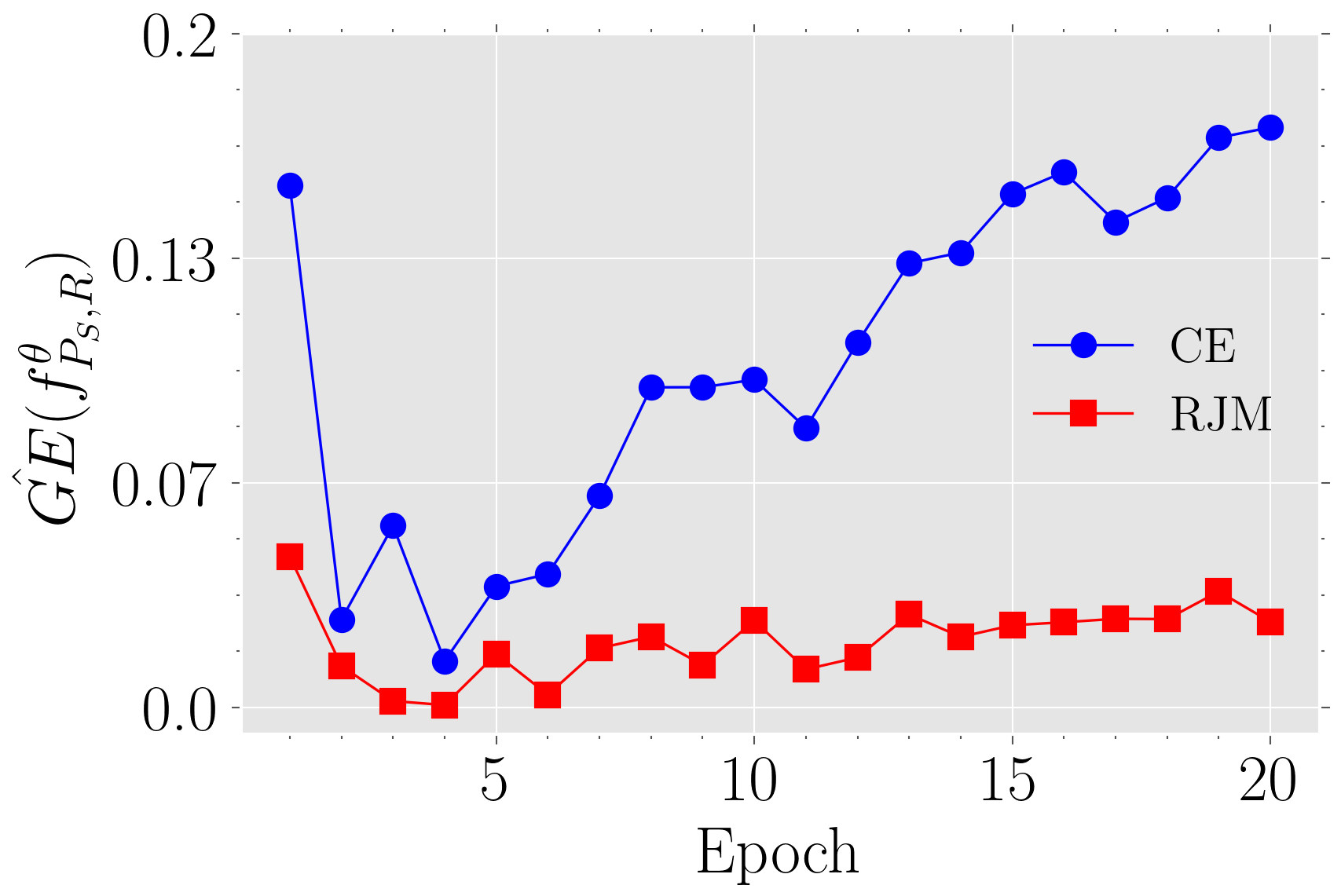}
			\caption{Generalization error estimate}
			\label{GE-ResNet-AdamW-CE-RJM}
		\end{subfigure}
		\hfill
		\begin{subfigure}[b]{0.3\textwidth}
			\includegraphics[height=3.33cm, width=5cm]{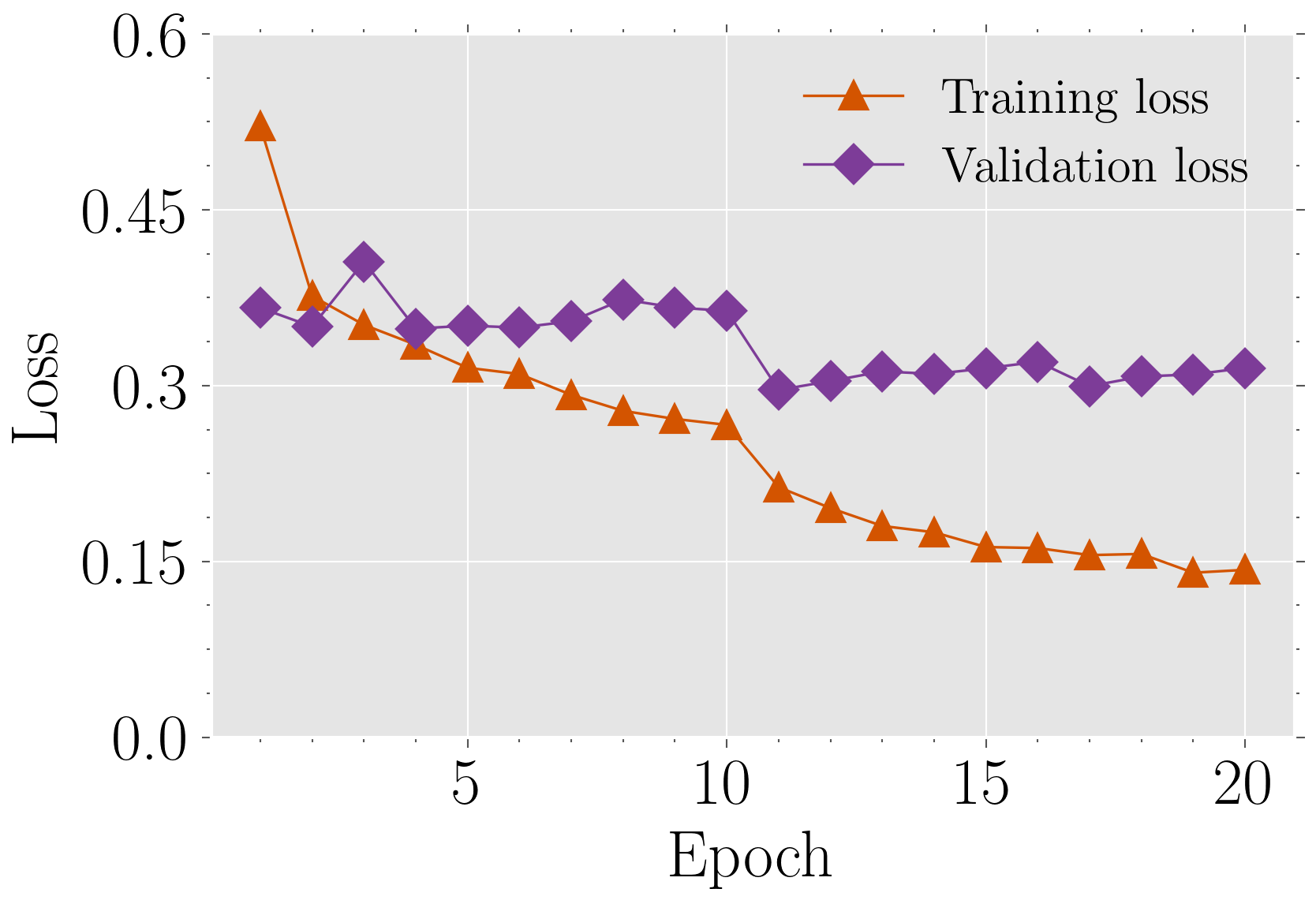}
			\caption{CE on the training and validation sets}
			\label{ResNet-AdamW-CE}
		\end{subfigure}
		\hfill
		\begin{subfigure}[b]{0.3\textwidth}
			\includegraphics[height=3.33cm, width=5cm]{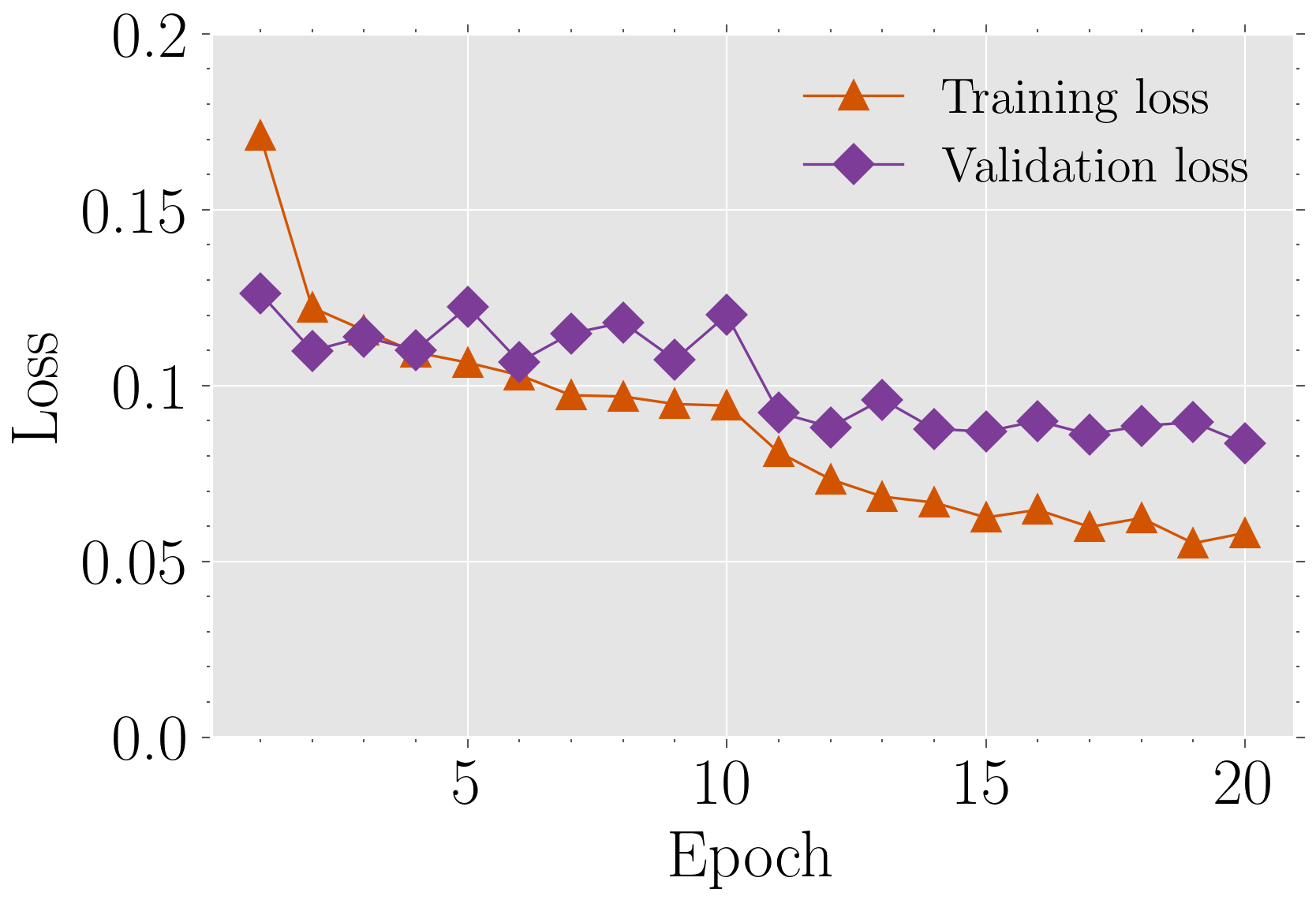}
			\caption{RJM on the training and validation sets}
			\label{ResNet-AdamW-RJM}
		\end{subfigure}
		\caption{Evaluation in terms of the generalization error estimate and loss values (Model: ResNet50, Optimizer: AdamW)}
	\end{center}
\end{figure}

\begin{figure}[h]
	\begin{center}
		\begin{subfigure}[b]{0.3\textwidth}
			\includegraphics[height=3.33cm, width=5cm]{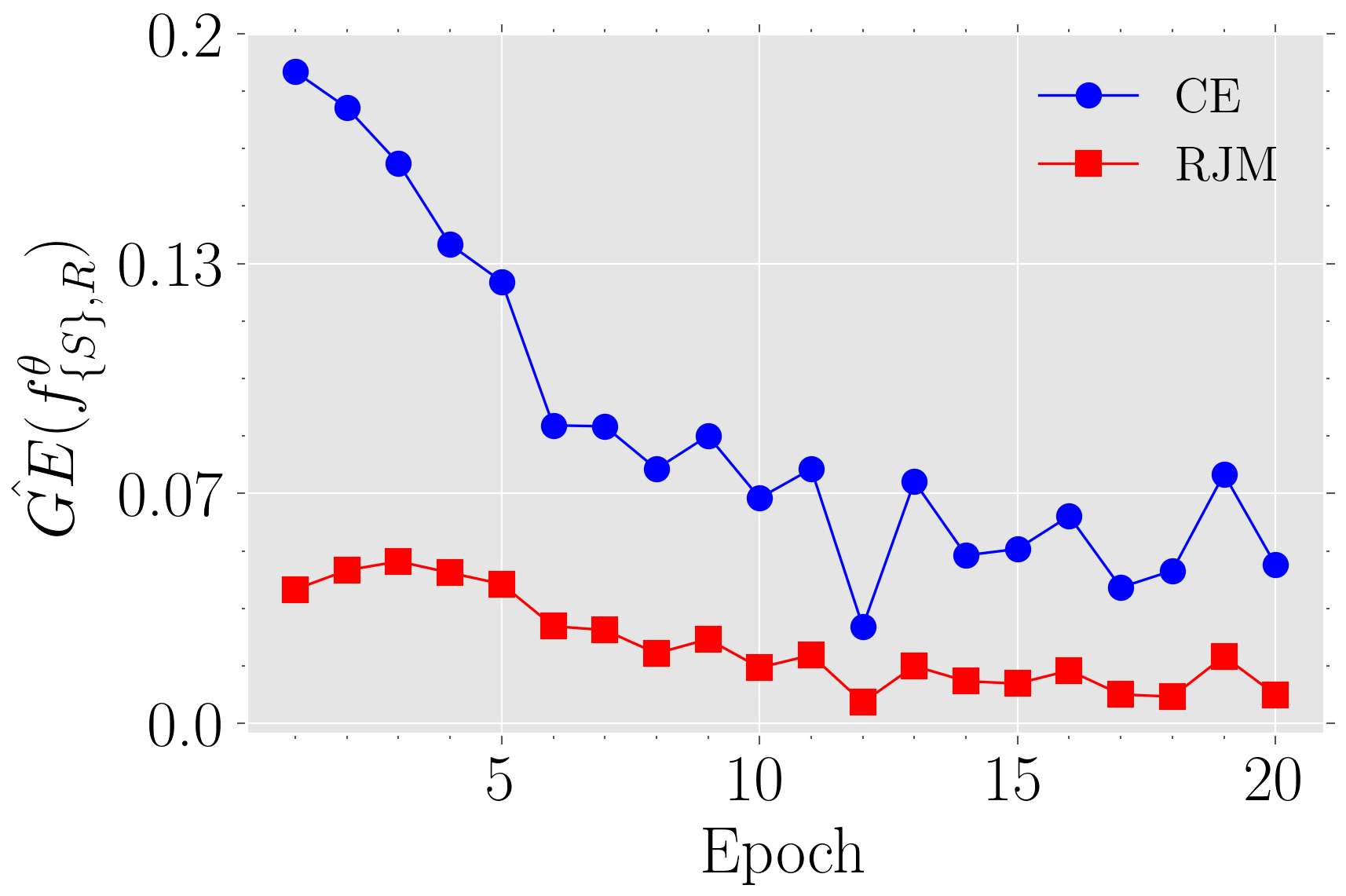}
			\caption{Generalization error estimate}
			\label{GE-VGG-SGD-CE-RJM}
		\end{subfigure}
		\hfill
		\begin{subfigure}[b]{0.3\textwidth}
			\includegraphics[height=3.33cm, width=5cm]{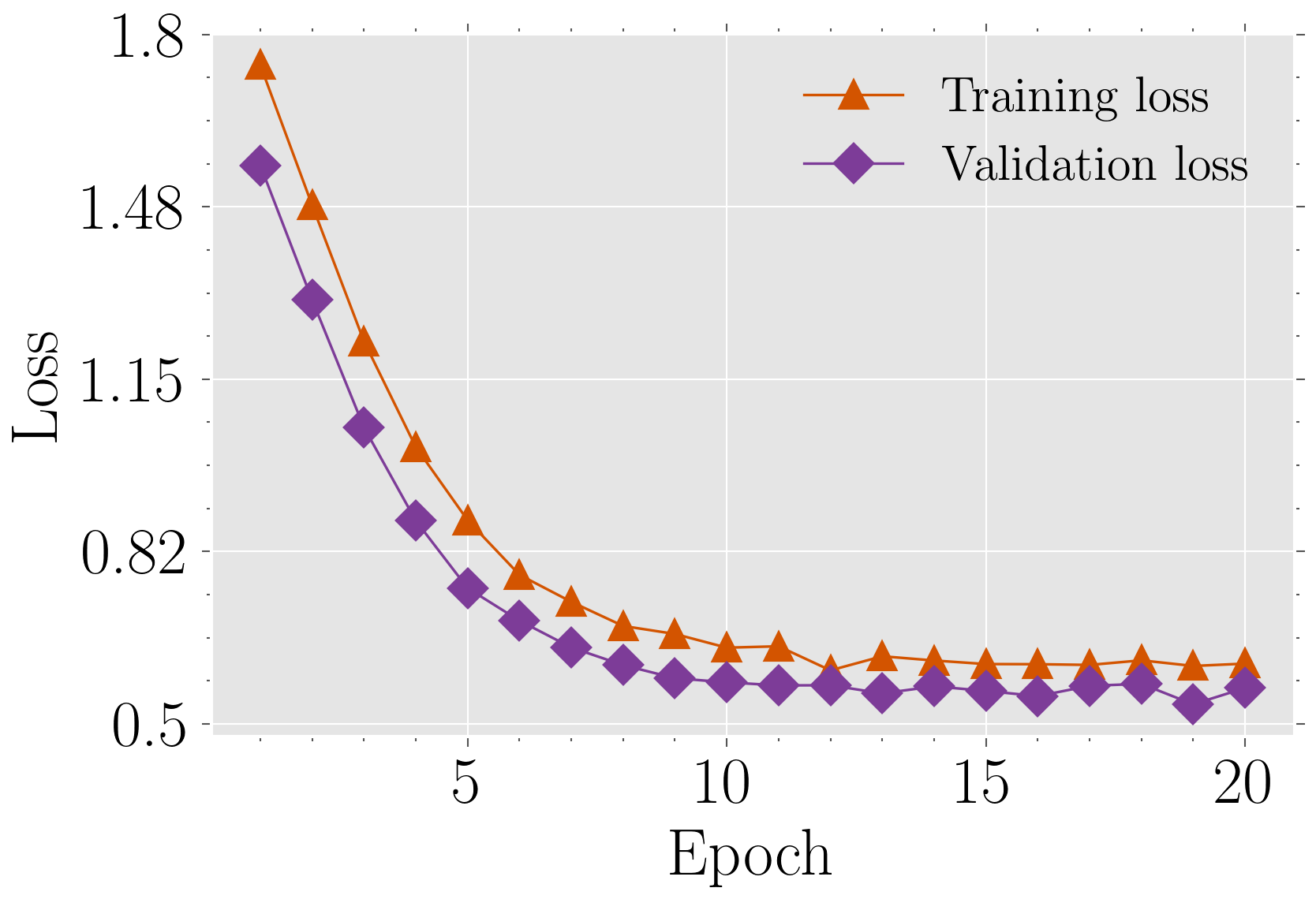}
			\caption{CE on the training and validation sets}
			\label{VGG-SGD-CE}
		\end{subfigure}
		\hfill
		\begin{subfigure}[b]{0.3\textwidth}
			\includegraphics[height=3.33cm, width=5cm]{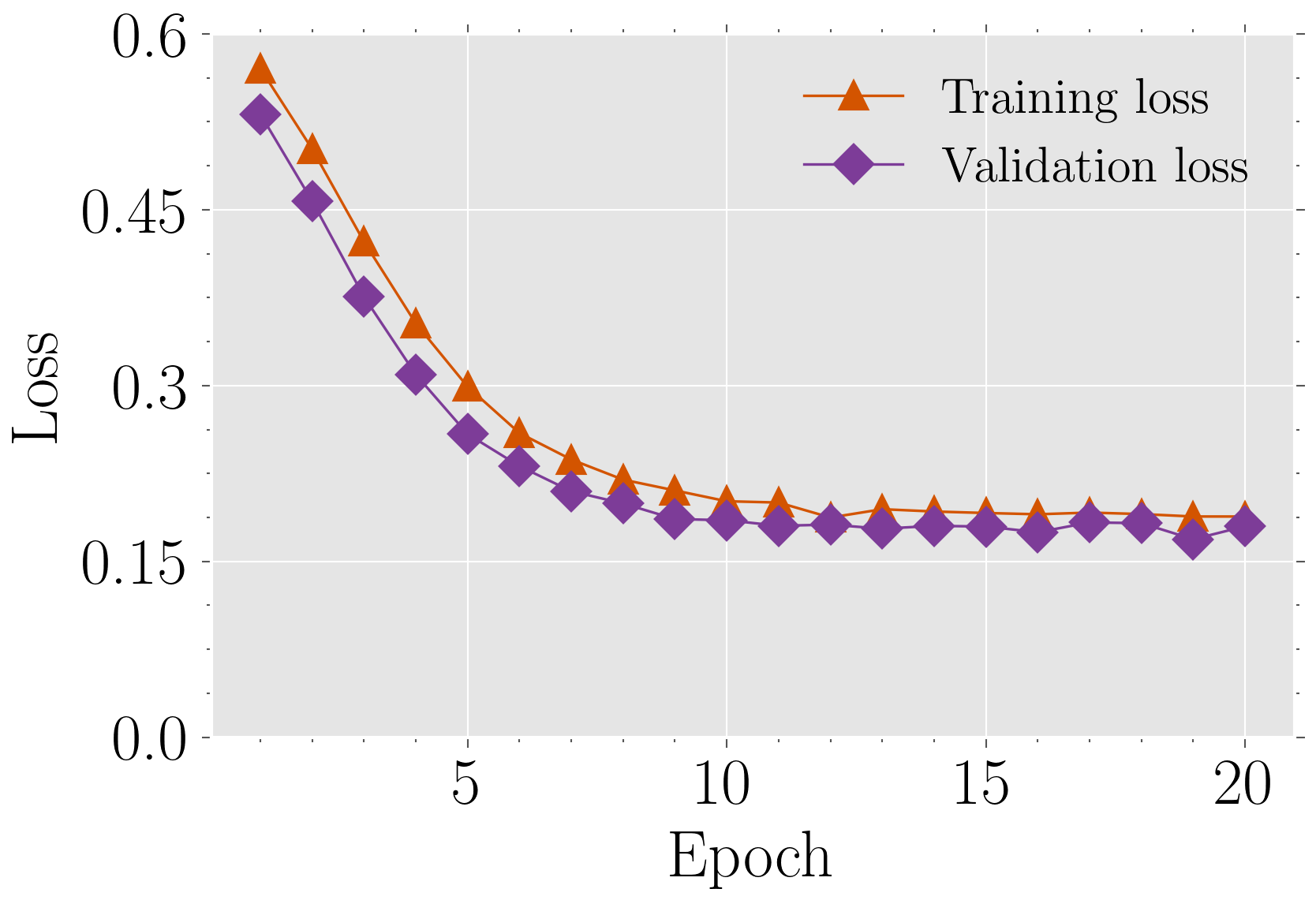}
			\caption{RJM on the training and validation sets}
			\label{VGG-SGD-RJM}
		\end{subfigure}
		\caption{Evaluation in terms of the generalization error estimate and loss values (Model: VGG16, Optimizer: SGD)}
	\end{center}
\end{figure}

As our first observation, we evaluate the generalization performance of our models by monitoring the generalization error estimate in every epoch:
\[
\hat{GE}(f^\theta_{P_S,R}) = |E_{train} (f^\theta_{P_S,R}) - E_{val} (f^\theta_{P_S,R}) |, 
\]
where $ E_{val} (f^\theta_{P_S,R}) $ is the average loss value on the validation set, and $ E_{train} (f^\theta_{P_S,R}) $ is the training error of the output model defined in Section \ref{prelim}. Figures \ref{GE-ResNet-Adam-CE-RJM}, \ref{GE-ResNet-AdamW-CE-RJM}, and \ref{GE-VGG-SGD-CE-RJM} show the effectiveness of RJM in preventing the over-fitting issue where the training and validation sets are Intel-Train and Intel-Val respectively. The plots demonstrate that training DNNs with RJM can reduce the generalization error and improve confidence in the output models. Comparing Figures \ref{VGG-SGD-CE} and \ref{VGG-SGD-RJM} to Figures \ref{ResNet-Adam-CE}, \ref{ResNet-Adam-RJM}, \ref{ResNet-AdamW-CE}, and \ref{ResNet-AdamW-RJM}, we realize that the output models obtained by SGD have been under-fitted. However, RJM was still effective in diminishing the generalization error. 

We also evaluate the generalization performance in terms of Accuracy and F1-score which are the specific metrics for classification tasks. The results are reported in Table \ref{results}. RJM increases the Accuracy and F1-score of the models trained by Adam, AdamW, and SGD on the test set. These metrics for models obtained by SGD are lower than others because it does not use the exponential moving average of the gradient vector to adapt it, leading to the under-fitting issue in our case. Furthermore, we report the training times in Table \ref{results} to show that the computation time of RJM is as much as CE.

\begin{table}[h]
  	\begin{center}
  		\caption{ Accuracy and F1-score on Intel-Test}
  		\begin{tabular}{|c|c|c|cc||c|}
  			\hline
  			Optimizer & Arch. & Loss function & Accuracy & F1-score & Training time \\
  			\hline \hline
  			\multirow{2}{*}{Adam} & \multirow{2}{4em}{ResNet50} & CE & $ 93.03 $ & $ 93.16 $ & $ 19\text{m} \ 19\text{s} $ \\  
  			&  & RJM & $ \mathbf{93.33} $ & $ \mathbf{93.44} $ & $ 19\text{m} \ 15\text{s} $ \\ \hline
  			\multirow{2}{*}{AdamW} & \multirow{2}{*}{ResNet50} & CE & $ 92.40 $ & $ 92.50 $ &  $ 20\text{m} \ 31\text{s} $ \\  
  			&  & RJM & $ \mathbf{93.27} $ & $ \mathbf{93.38} $ & $ 20\text{m} \ 25\text{s} $ \\ \hline
  			\multirow{2}{*}{SGD} & \multirow{2}{*}{VGG16} & CE & $ 79.55 $ & $ 79.59 $ & $ 25\text{m} \ 35\text{s} $ \\  
  			&  & RJM & $\mathbf{79.65}$ & $\mathbf{79.79}$ & $ 25\text{m} \ 36\text{s} $ \\ \hline
  		\end{tabular}
  		\label{results}
  	\end{center}
\end{table}

\subsection{Node Classification}
\subsubsection{Problem Formulation}
Node classification is a single-label learning problem in the graph learning domain. Consider a graph $\mathcal{G} = (\mathcal{V},\mathcal{E})$ where $ \mathcal{V} $ is the node set and $\mathcal{E} \subseteq \mathcal{V} \times \mathcal{V}$ is the set of edges. $ \mathcal{E} $ indicates the connection between the nodes which is used to pass messages throughout the graph in the learning process. Let $ (v, c) $ be a sample of this problem such that $v \in \mathcal{V}$ and $c \in \lbrace 1,2,\ldots,C \rbrace$. Given a mapping function $f^\theta_{\mathcal{V}, \mathcal{E}}$, the predicted probability vector of size $C$ corresponding to $v$ is $\hat{\mathrm{y}} = f^\theta_{\mathcal{V}, \mathcal{E}}(v)$. The class with the highest probability is the predicted label.

In recent years, graph neural networks (GNNs) have been widely exploited to solve this problem. GNNs learn representations of nodes and predict the corresponding label to the input from end to end. Two major components of these architectures in each layer are message transformation and aggregation which create computation trees for each node. We study pioneering models GCN \cite{kipf2016semi}, GraphSAGE \cite{hamilton2017inductive}, and GAT \cite{velivckovic2017graph} to assess the new loss function in the graph learning context.
\subsubsection{Dataset and Settings}
We use a specific version of the CiteSeer dataset \cite{citeseer2016} containing $3327$ articles, classified into $6$ classes, and form a citation network. The number of training, validation, and test nodes are $120$, $500$, and $1000$ respectively. Other nodes are isolated. Due to the small number of training nodes, we can evaluate the models in the over-fitting issue appropriately. There are $9104$ edges in this graph and all nodes have $3703$ features.  

Now we explain our settings for the GNNs. GCN has two hidden layers. Each GraphSAGE and GAT has one hidden layer. The number of hidden channels of GCN, GraphSAGE, and GAT are $8$, $16$, and $64$ respectively. We only use Adam to train the models because based on our observations, it is the most widely used optimizer in graph learning. The learning rate is set to $0.001$, as suggested by \cite{kingma2014adam}. The values of $\beta_1$ and $\beta_2$ are set to $0.9$ and $0.999$, respectively. We train the GCN model in $100$ epochs. GraphSAGE and GAT are trained in $200$ epochs. We save the models at the epoch they have the minimum validation loss.
\subsubsection{Evaluation}
\begin{figure}[h]
	\begin{center}
		\begin{subfigure}[b]{0.3\textwidth}
			\includegraphics[height=3.33cm, width=5cm]{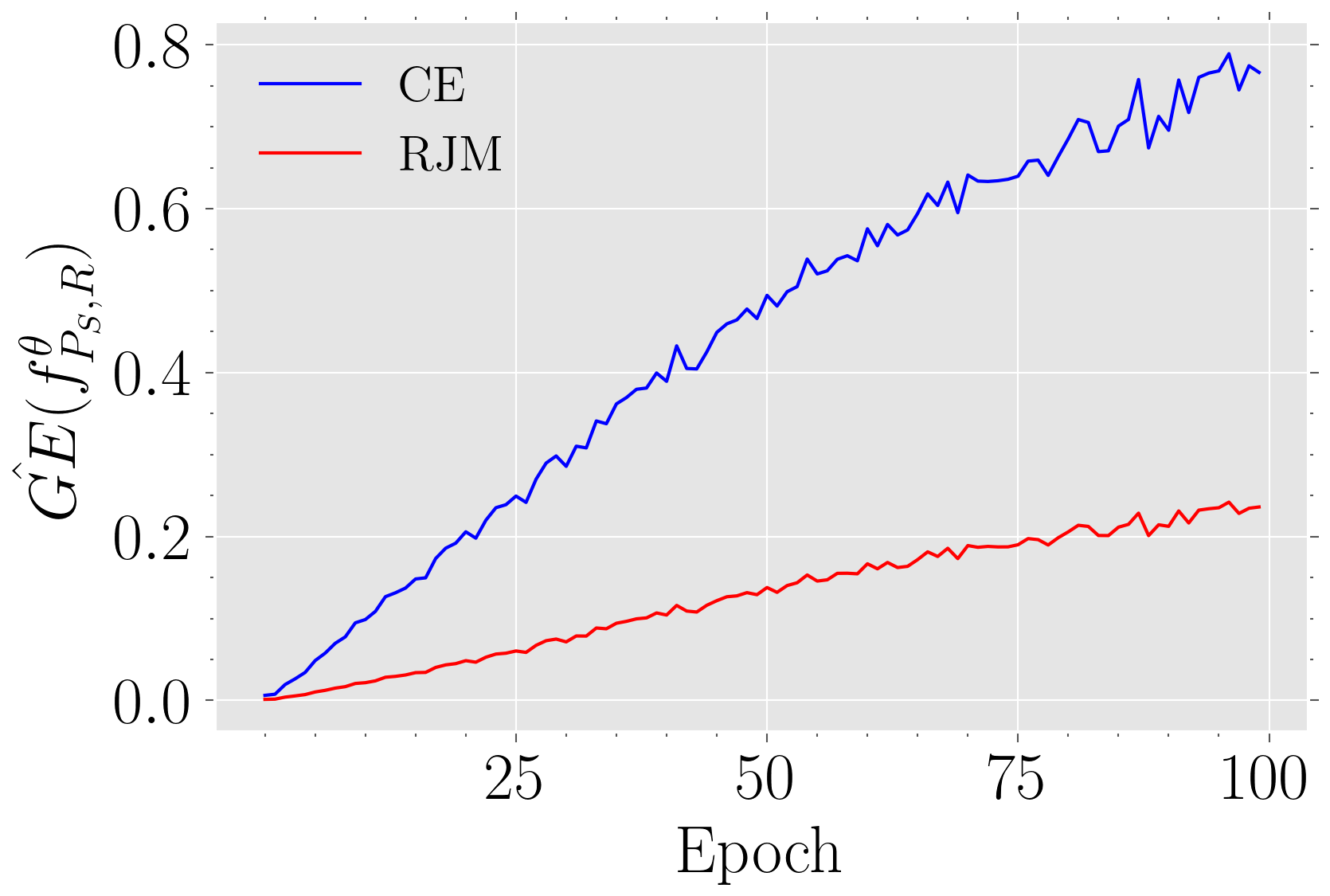}
			\caption{Generalization error estimate}
			\label{GE-GCN-Adam-CE-RJM}
		\end{subfigure}
		\hfill
		\begin{subfigure}[b]{0.3\textwidth}
			\includegraphics[height=3.33cm, width=5cm]{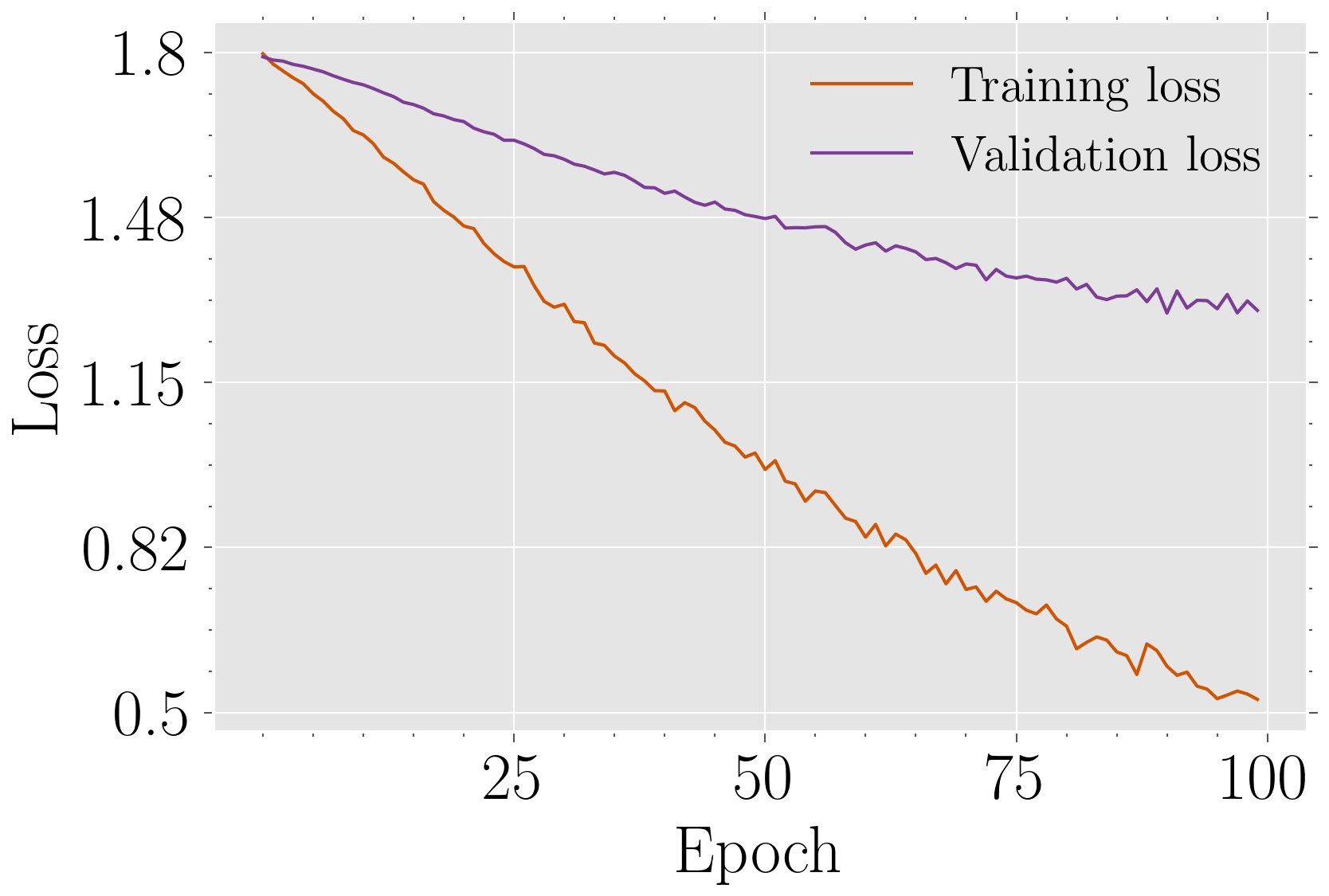}
			\caption{CE on the training and validation sets}
			\label{GCN-Adam-CE}
		\end{subfigure}
		\hfill
		\begin{subfigure}[b]{0.3\textwidth}
			\includegraphics[height=3.33cm, width=5cm]{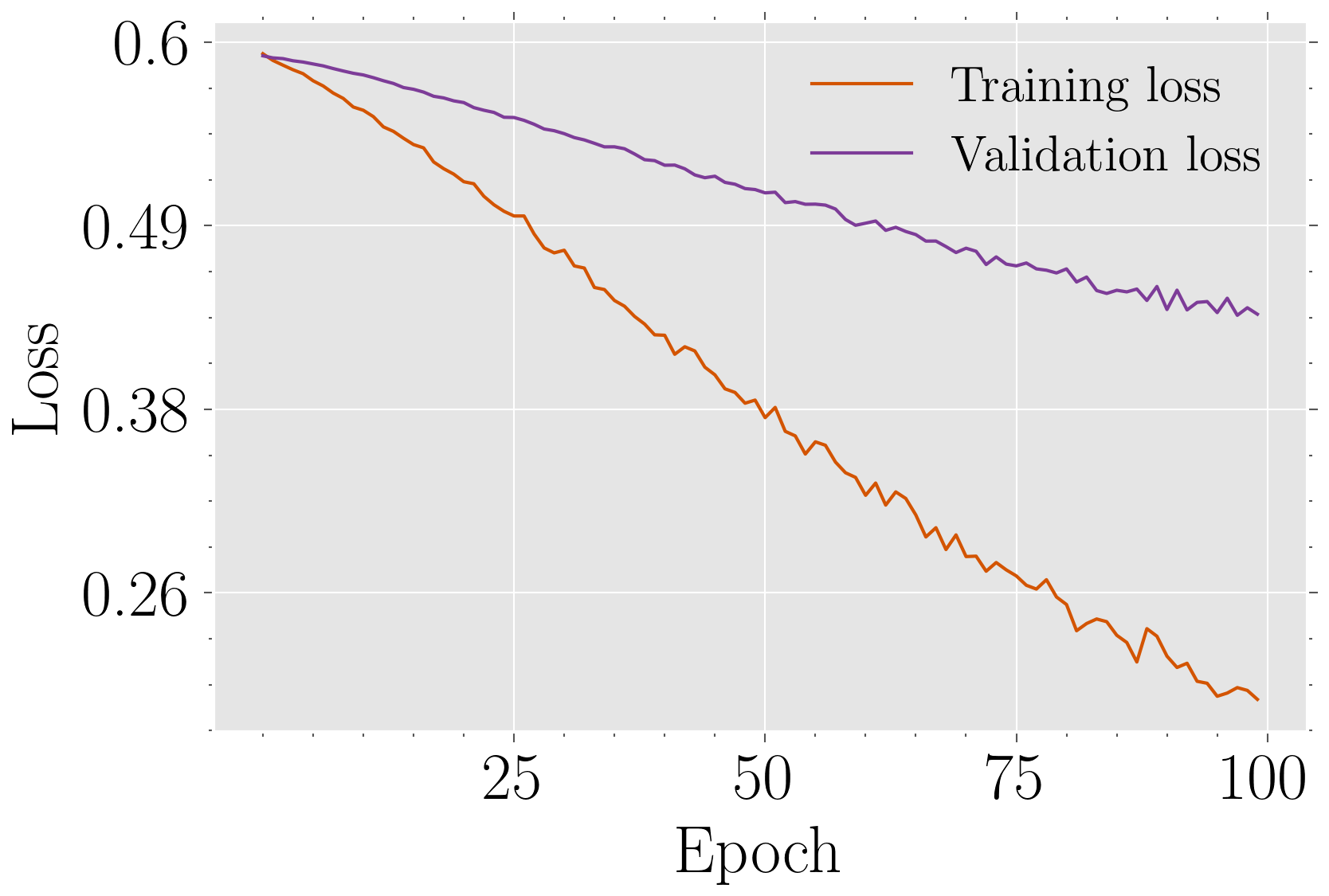}
			\caption{RJM on the training and validation sets}
			\label{GCN-Adam-RJM}
		\end{subfigure}
		\caption{Evaluation in terms of the generalization error estimate and loss values (Model: GCN)}
	\end{center}
\end{figure}

\begin{figure}[h]
	\begin{center}
		\begin{subfigure}[b]{0.3\textwidth}
			\includegraphics[height=3.33cm, width=5cm]{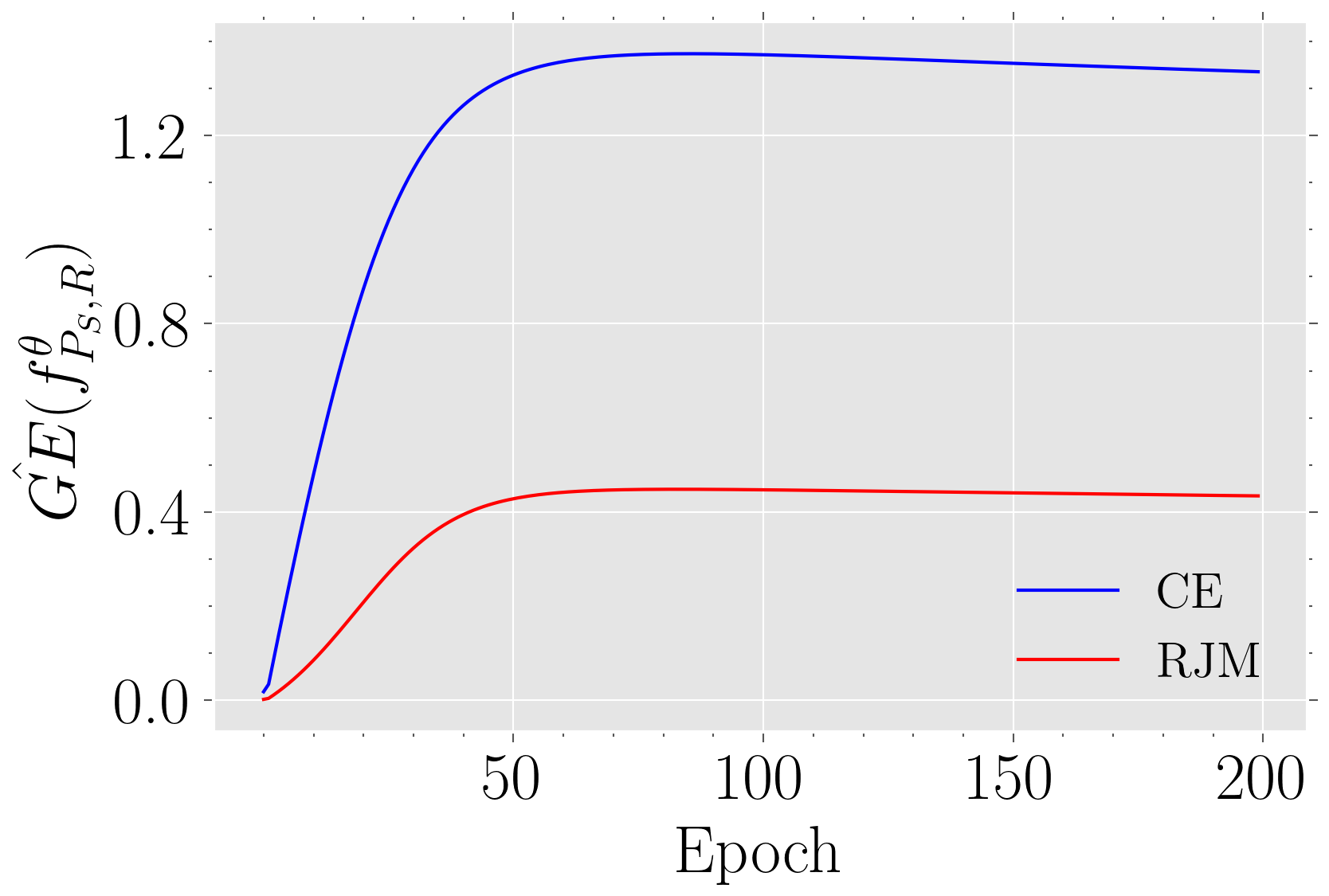}
			\caption{Generalization error estimate}
			\label{GE-SAGE-Adam-CE-RJM}
		\end{subfigure}
		\hfill
		\begin{subfigure}[b]{0.3\textwidth}
			\includegraphics[height=3.33cm, width=5cm]{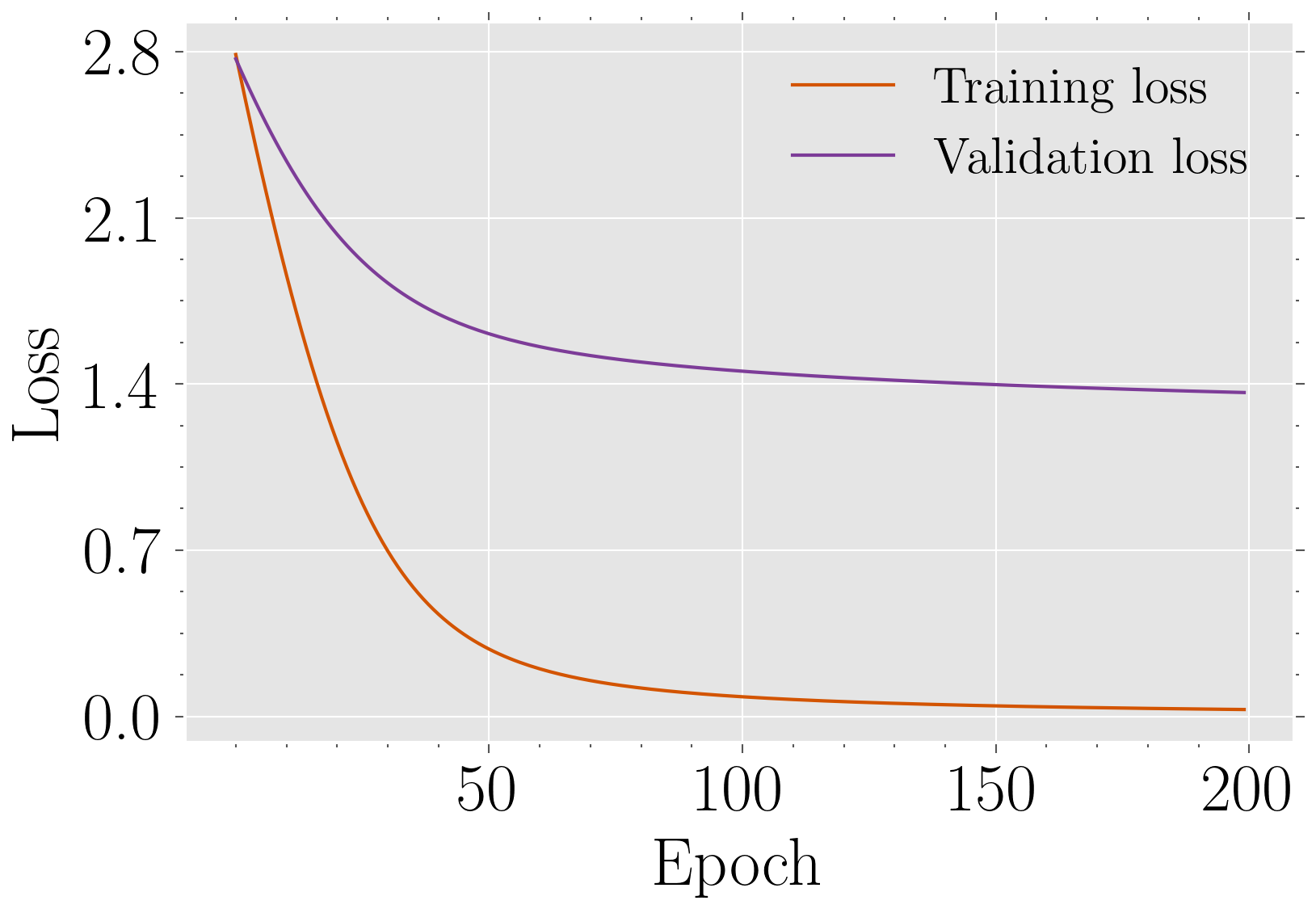}
			\caption{CE on the training and validation set}
			\label{GCN-SAGE-CE}
		\end{subfigure}
		\hfill
		\begin{subfigure}[b]{0.3\textwidth}
			\includegraphics[height=3.33cm, width=5cm]{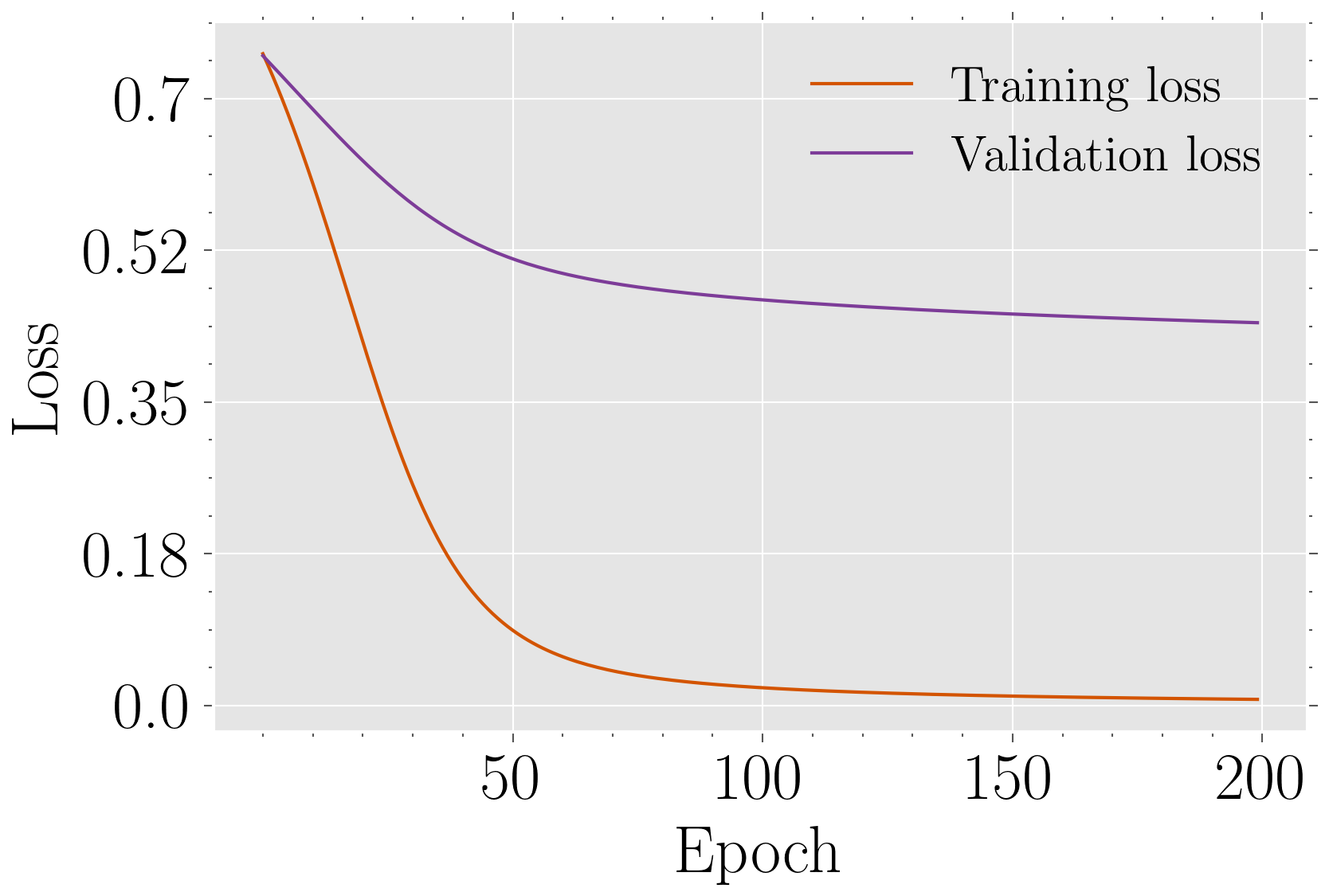}
			\caption{RJM on the training and validation sets}
			\label{GCN-SAGE-RJM}
		\end{subfigure}
		\caption{Evaluation in terms of the generalization error estimate and loss values (Model: GraphSAGE)}
	\end{center}
\end{figure}

\begin{figure}[h]
	\begin{center}
		\begin{subfigure}[b]{0.3\textwidth}
			\includegraphics[height=3.33cm, width=5cm]{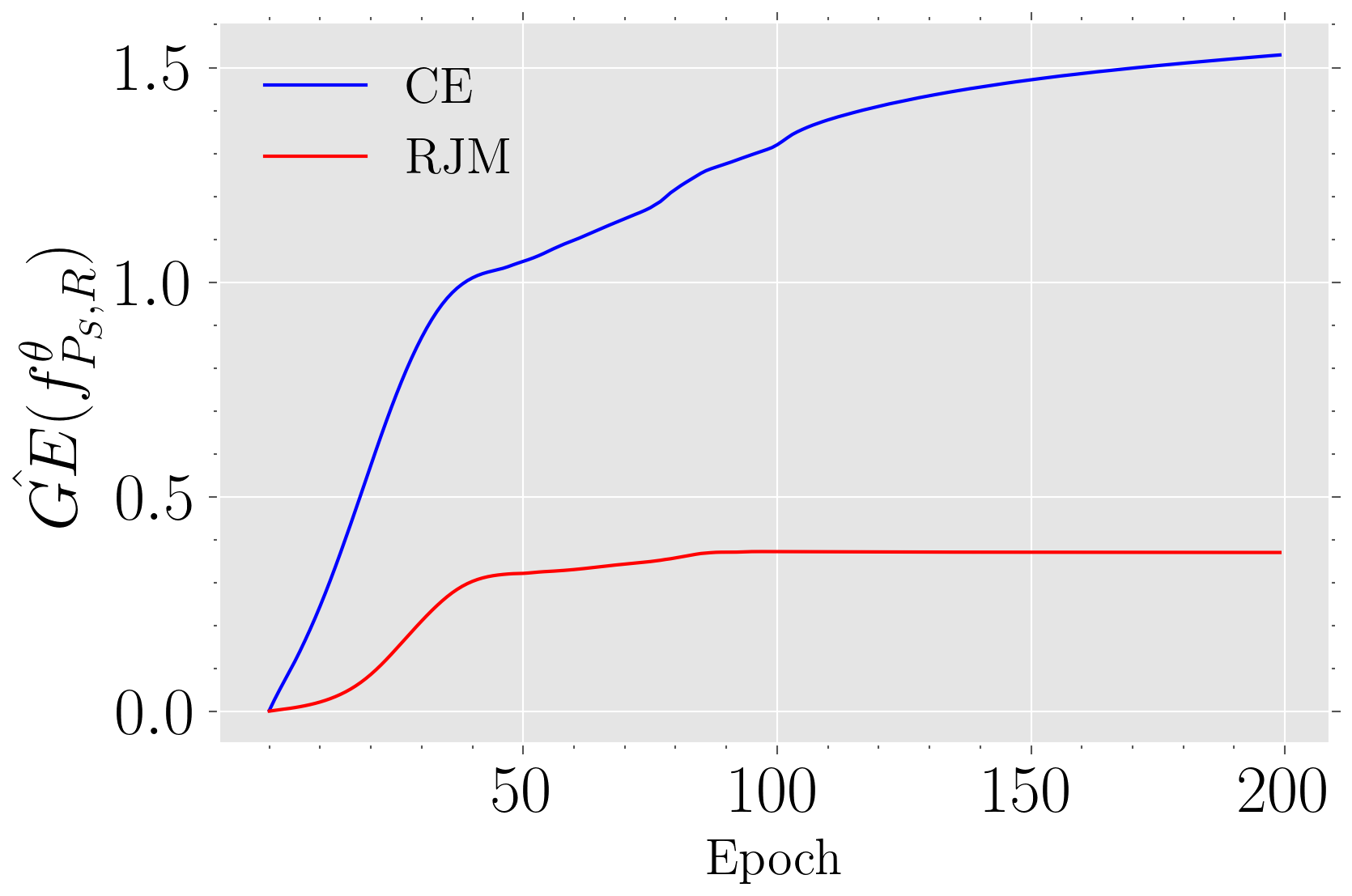}
			\caption{Generalization error estimate}
			\label{GE-GAT-Adam-CE-RJM}
		\end{subfigure}
		\hfill
		\begin{subfigure}[b]{0.3\textwidth}
			\includegraphics[height=3.33cm, width=5cm]{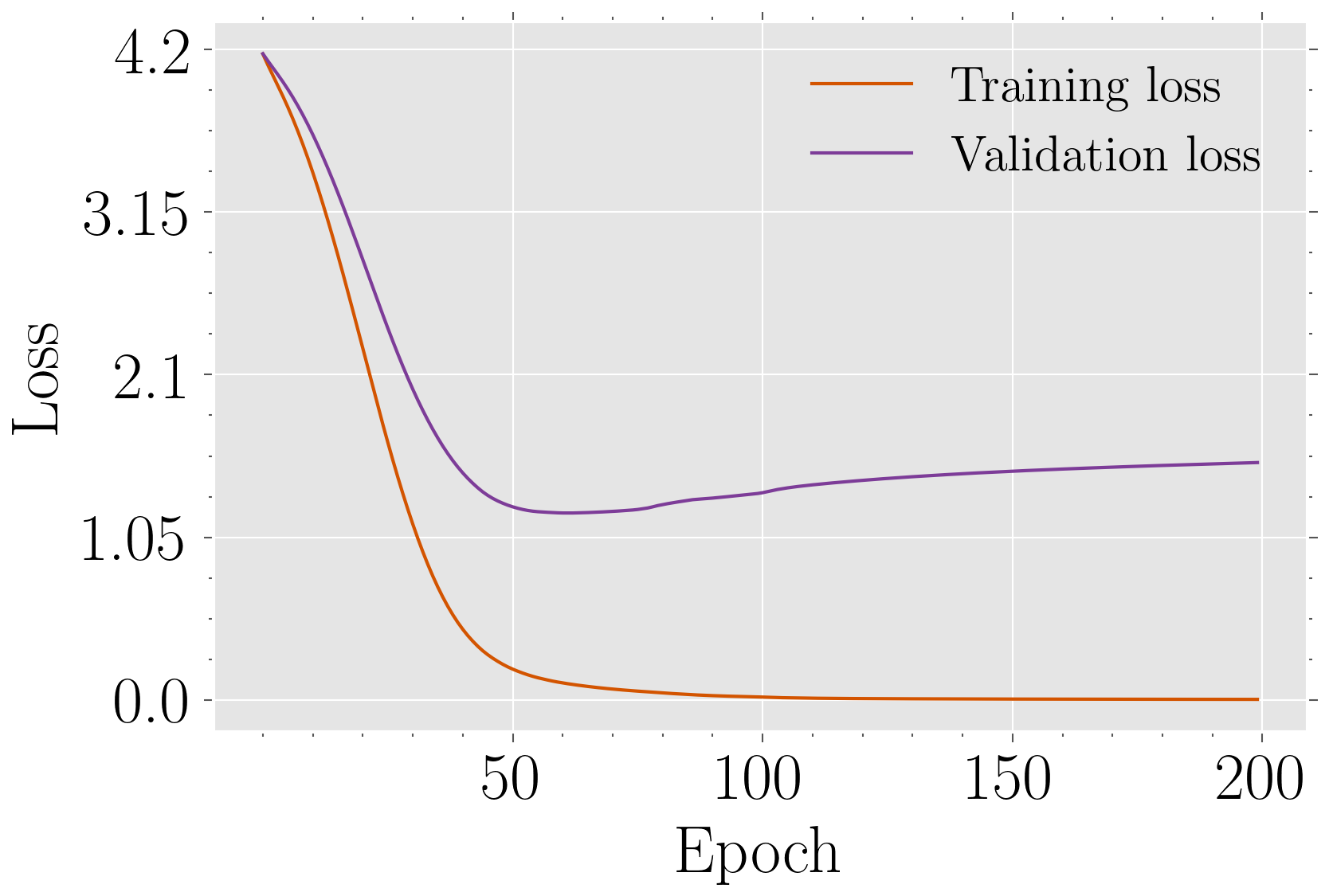}
			\caption{CE on the training and validation set}
			\label{GCN-GAT-CE}
		\end{subfigure}
		\hfill
		\begin{subfigure}[b]{0.3\textwidth}
			\includegraphics[height=3.33cm, width=5cm]{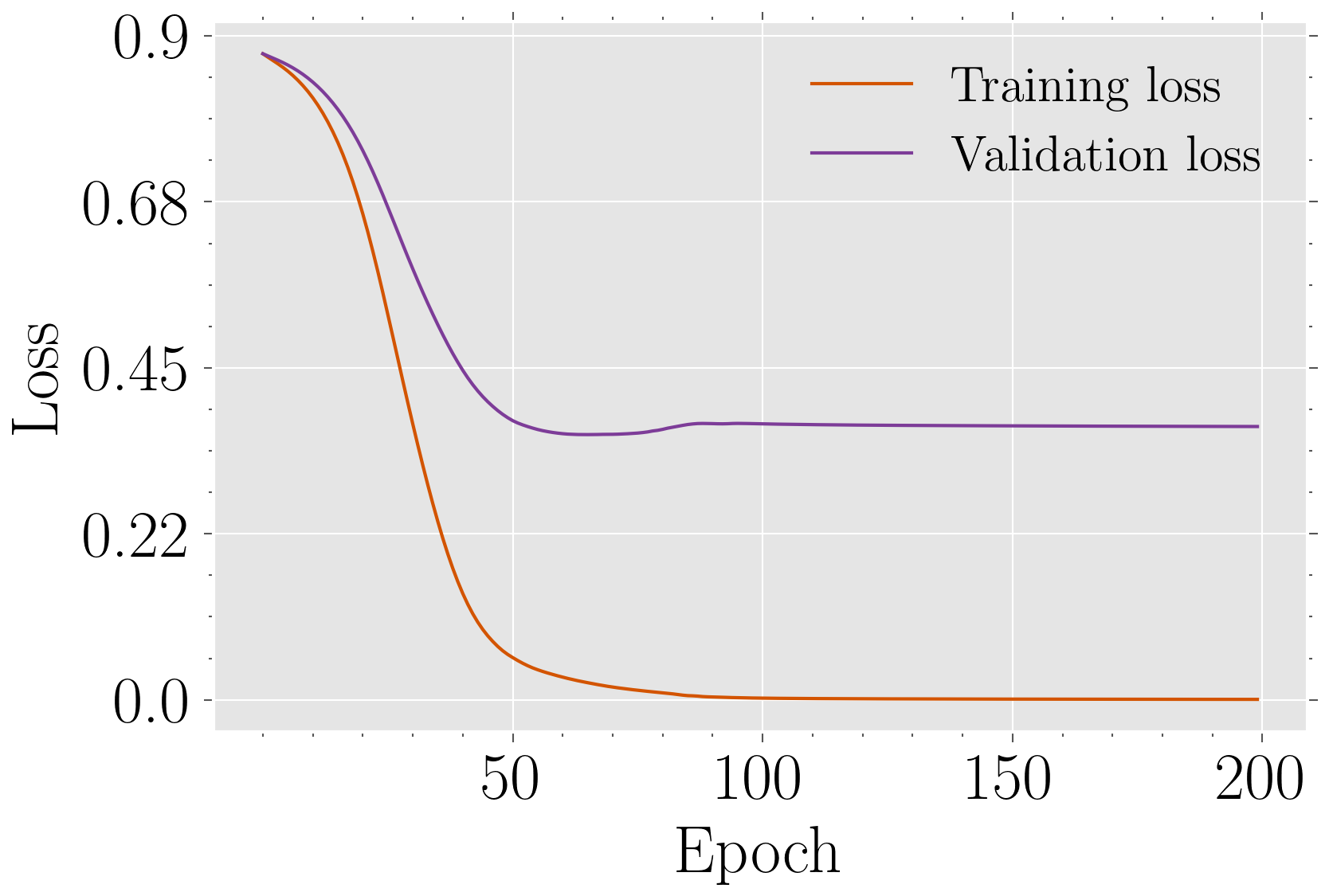}
			\caption{RJM on the training and validation sets}
			\label{GCN-GAT-RJM}
		\end{subfigure}
		\caption{Evaluation in terms of the generalization error estimate and loss values (Model: GAT)}
	\end{center}
\end{figure}

As shown in Figures \ref{GE-GCN-Adam-CE-RJM}, \ref{GE-SAGE-Adam-CE-RJM}, and \ref{GE-GAT-Adam-CE-RJM}, the over-fitting issue is alleviated for the models trained using RJM. The figures illustrate that the generalization error estimate of all three models becomes closer to zero significantly when the loss function is RJM. 

The Accuracy and F1-score metrics on Test and Validation sets are reported in Table \ref{results3}. The results show that RJM can perform better than CE in the domain of node classification. GAT models outperform the others because, in this architecture, attention parameters are used to aggregate messages. The best model is GAT, trained using RJM. The number of parameters of the models and training samples is very low. Hence, the models take a negligible amount of time in the training phase, having no value to report.

\begin{table}[h]
	\begin{center}
		\caption{Accuracy and F1-score on Validation and Test Sets}
		\begin{tabular}{|c|c|cc|cc|}
			\hline
			Arch. & Loss function & Validation Accuracy & Validation F1-score & Test Accuracy & Test F1-score\\ 
			\hline \hline
			\multirow{2}{*}{GCN} & CE & $ 62.00 $ & $ 57.16 $ & $ 60.70 $ & $ 57.73 $ \\  			
			& RJM & $ \mathbf{63.80} $ & $ \mathbf{58.84} $ & $ \mathbf{61.30} $ & $ \mathbf{58.21} $ \\ 
			\hline
			\multirow{2}{*}{GraphSAGE} & CE & $ 60.60 $ & $ 57.76 $ & $ 58.80 $ & $ 55.99 $  \\  
			& RJM & $ \mathbf{61.80} $ & $ \mathbf{59.03} $ & $ \mathbf{59.20} $ & $ \mathbf{56.33} $ \\ 
			\hline
			\multirow{2}{*}{GAT} & CE & $ 62.40 $ & $ 60.20 $ & $ 62.10 $ & $ 59.72 $ \\  
			& RJM & $ \mathbf{63.00} $ & $ \mathbf{60.96} $ & $ \mathbf{62.40} $ & $ \mathbf{59.91} $ \\ 
			\hline
		\end{tabular}
		\label{results3}
	\end{center}
\end{table}

\section{Conclusion and Future Work}
In this paper, we proposed the RJM loss function to diminish the generalization error of DNNs in classification tasks using generalization bounds previously found under the uniform stability approach, distinguishing the role of loss functions in improving the generalization of deep learning models. Comparing RJM to CE in image and node classification problems, we conclude that RJM reduces over-fitting and increases the value of Accuracy and F1-score.

RJM can also prevent the over-fitting issue of probabilistic models in other machine learning tasks (e.g. image segmentation, part of speech tagging, named entity recognition, recommender systems). Additionally, RJM is applicable in the multi-label learning framework i.e. single-label classification on each feature. Note that Lemma \ref{lemma1} can be utilized to create different novel loss functions in classification tasks. 

As a future work, we suggest upper-bounding the difference between the true and training values of specific evaluation metrics for classification models (e.g. Accuracy and F1-score) in terms of loss function characteristics and hyper-parameters of a deep learning problem because the relationship between loss functions and these metrics is not distinguished by generalization bounds directly.






\bibliographystyle{ieeetr}
\bibliography{ref}

\end{document}